%% file: main.tex
\pdfoutput=1

\documentclass[11pt]{article}

\usepackage{EMNLP2022}

\usepackage{times}
\usepackage{latexsym}
\usepackage{amsfonts}
\usepackage{amsmath}
\usepackage{pifont}
\usepackage[normalem]{ulem}
\usepackage{pgfplots}
\usetikzlibrary{pgfplots.colorbrewer}
\usepackage{svg}
\usepackage{graphicx}
\usepackage{subfig}

\usepackage{booktabs}
\usepackage{multirow}
\usepackage{wrapfig}
\usepackage{diagbox}
\usepackage{makecell}

\usepackage[T1]{fontenc}

\usepackage[utf8]{inputenc}

\usepackage{microtype}

\usepackage{inconsolata}

\newcommand\our{\textsc{XY-LENT}}

\newcommand\baseline{\textsc{XLM-E}}
\DeclareMathOperator{\Tr}{Tr}

\title{Beyond English-Centric Bitexts for Better Multilingual Language Representation Learning}

\author{{ Barun Patra\Thanks{ Equal Contribution.} }, { Saksham Singhal\footnotemark[1] }, { Shaohan Huang\footnotemark[1] }, \\ \textbf{Zewen Chi}, \textbf{Li Dong}, \textbf{Furu Wei}, \textbf{Vishrav Chaudhary}, \textbf{Xia Song} \\
    Microsoft \\
    {\tt \{bapatra, saksingh, shaohanh, v-zewenchi, lidong1,} \\
     \tt{ fuwei, vchaudhary, xiaso\}@microsoft.com}
}

\pgfplotsset{compat=1.17} 
\begin{document}
\maketitle
\begin{abstract}
In this paper, we elaborate upon recipes for building multilingual representation models that are not only competitive with existing state-of-the-art models but are also more parameter efficient, thereby promoting better adoption in resource-constrained scenarios and practical applications. We show that going beyond English-centric bitexts, coupled with a novel sampling strategy aimed at reducing under-utilization of training data, substantially boosts performance across model sizes for both Electra and MLM pre-training objectives. We introduce \textit{\textbf{\our{}}}: \textbf{X-Y} bitext enhanced \textbf{L}anguage \textbf{EN}codings using \textbf{T}ransformers which not only achieves state-of-the-art performance over 5 cross-lingual tasks within all model size bands, is also competitive across bands. Our \our{}\textsubscript{XL} variant outperforms XLM-R\textsubscript{XXL} and exhibits competitive performance with mT5\textsubscript{XXL} while being 5x and 6x smaller respectively. We then show that our proposed method helps ameliorate the curse of multilinguality, with the \our{}\textsubscript{XL} achieving 99.3\% GLUE performance and 98.5\% SQuAD 2.0 performance compared to a SoTA English only model in the same size band. We then analyze our models performance on extremely low resource languages and posit that scaling alone may not be sufficient for improving the performance in this scenario.

\end{abstract}

\input{1.introduction}
\input{1.related_work}
\input{2.data-scaling}
\input{3.model-scaling}
\input{4.experiment-details}
\input{5.results}

\input{6.conclusion}

\input{7.limitations}

\bibliography{anthology,custom}
\bibliographystyle{acl_natbib}

\appendix

\input{Appendix}

\end{document}

%% file: 1.introduction.tex
\section{Introduction}
\pgfplotsset{compat = newest}
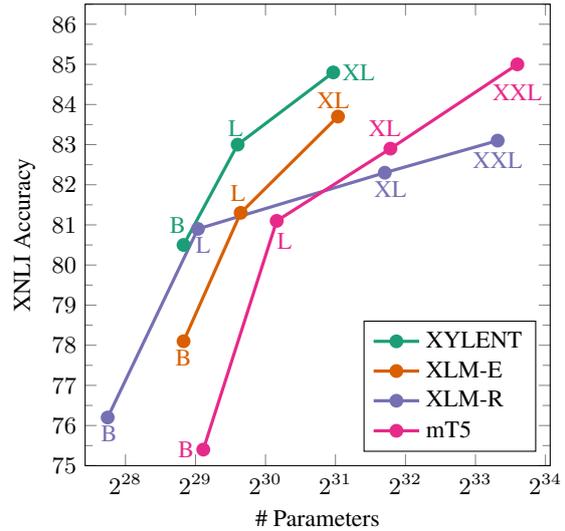
\begin{figure}[ht]
    \footnotesize
    \pgfplotstableread{data/xnli_comp.dat}{\table}
    \begin{tikzpicture}
    \begin{axis}[
        xmin = 180000000, xmax = 18000000000,
        ymin = 75, ymax = 86.5,
        x tick label style={
		/pgf/number format/1000 sep=},
        ytick distance = 1,
        xmode=log,
        log basis x={2},
        xtick={268435456,536870912,1073741824,2147483648,4294967296
,8589934592,17179869184},
        ylabel={XNLI Accuracy},
        xlabel={\# Parameters},
        minor tick num = 1,
        major grid style = {lightgray},
        minor grid style = {lightgray!25},
        width = \columnwidth,
        height = \columnwidth,
        legend cell align = {left},
        legend pos = south east,
        cycle list/Dark2,
        every axis plot/.append style={very thick}
    ]
    \addplot+[mark=*] table [x ={x}, y = {y4}] {\table};
    
    \addplot+[mark=*] table [x = {x}, y = {y3}] {\table};
    
    \addplot+[mark=*] table [x = {x}, y = {y1}] {\table};
     
    \addplot+[mark=*] table [x ={x}, y = {y2}] {\table};
    \node[text=xylent] at (axis cs:450000000,81) {B};
    \node[text=xlmr] at (axis cs:228000000,75.8) {B};
    \node[text=xlme] at (axis cs:477000000,77.7) {B};
    \node[text=mt5] at (axis cs:490000000,75.4) {B};
    
    \node[text=xylent] at (axis cs:800000000,83.4) {L};
    \node[text=xlmr] at (axis cs:580000000,80.5) {L};
    \node[text=xlme] at (axis cs:820000000,81.8) {L};
    \node[text=mt5] at (axis cs:1300000000,80.6) {L};
    
    \node[text=xylent] at (axis cs:2700000000,84.8) {XL};
    \node[text=xlmr] at (axis cs:3700000000,81.9) {XL};
    \node[text=xlme] at (axis cs:2100000000,84.1) {XL};
    \node[text=mt5] at (axis cs:3500000000,83.4) {XL};

    \node[text=xlmr] at (axis cs:10700000000,82.6) {XXL};
    \node[text=mt5] at (axis cs:13000000000,84.3) {XXL};
     
    \legend{
        XYLENT, 
        XLM-E,
        XLM-R,
        mT5
    }
     
    \end{axis}
     
    \end{tikzpicture}
    \caption{The proposed XY-LENT model (green line) achieves SoTA performance within all band sizes and is competitive performance across larger model-size bands. The parameter efficiency of XY-LENT\textsubscript{XL} particularly stands out, outperforming XLM-R\textsubscript{XXL} and being competitive with mT5\textsubscript{XXL} while being 5x and 6x smaller than them respectively. We also present the performance of XLM-E which used as a baseline in this paper. }
    \label{fig:size_comparison_on_xnli}
\end{figure}

Recent advancements in Natural Language Processing (NLP) have been a direct consequence of leveraging foundational models \citep{foundation-models}, pretrained on a large text corpora in a self-supervised fashion. This has also been the case for multilingual NLP where pre-trained models like multilingual BERT (mBERT) \citep{multilingualBERTmd, bert}, XLM \citep{conneau2019cross}, XLM-Roberta \citep{xlm-roberta}, XLM-Electra \citep{xlm-e} and mT5 \citep{mT5} have all shown non-trivial performance gains, especially in the setup of zero-shot transfer, and have been the work-horse for a diverse number of multilingual tasks. Given their ubiquitous applicability in zero-shot downstream scenarios, improving the quality and enabling their usage in resource-constrained applications is also an important vein of research which we explore in this paper.

A source of improvement for these models has been leveraging bitext data for better representation learning \citep{conneau2019cross, xlm-e}. Most prior work, however, has focused on leveraging English-centric (\textit{EN-X}) bitext data. Contemporaneously, the related area of Massively Multilingual Machine Translation (a single model for translating between different pairs of languages, eg: \citet{aharoni-etal-2019-massively,zhang-etal-2020-improving,m2m100}) has shown tremendous progress, with \citet{m2m100} showing that a crucial aspect of this improvement has been moving beyond \textit{EN-X} parallel corpora and leveraging web-based mined \textit{X-Y} bitexts spanning 1000s of translation directions \citep{wikimatrix,multiccaligned,ccmatrix}. This makes a compelling case to explore if leveraging \textit{X-Y} bitexts can also improve multilingual representation learning. 

In this work, we introduce \textit{\textbf{\our{}}} (\textit{pronounced as "\textbf{Excellent}"}): \textbf{\textit{X-Y}} bitext enhanced \textbf{L}anguage \textbf{EN}codings using \textbf{T}ransformers. We first identify problems with using the commonly used sampling strategy proposed in \citet{m2m100}, showing that it induces sparse sampling distributions leading to under-utilization of data, and thus propose a novel strategy to mitigate this issue (\S \ref{method:sampling}). We then propose leveraging \textit{X-Y} bitexts in conjunction with the improved sampling strategy, as well as a VoCAP \citep{vocap} style sentencepiece vocabulary re-construction for improving multilingual representation learning (\S \ref{sec:dataset}). We show that our proposed method improves performance across all model size bands (\S \ref{result}). Additionally, we show that the performance gains hold for both Masked Language Models (MLM) and ELECTRA style models, affording an almost 12x speedup in training for the former (\S \ref{sec:ablation}). We systematically analyse the impact of model scaling with respect to the curse of multilinguality \cite{xlm-roberta} to observe that the gap between current English only SoTA models and multilingual models can be considerably reduced (\S \ref{result:en-centric}). Our analysis reveals that \our{} improves performance across language families (\S \ref{sec:language-families}) and helps reduce the cross-lingual transfer gap in multilingual tasks (\S \ref{sec:crosslingual-transfer-gap}). We then demonstrate that the training dynamics of such models can be used to better understand the underlying datasets and use it to find interesting defects in them (\S \ref{result:dataset}). Finally, we show some limitations of such multilingual representational models vis-à-vis extremely low resource languages, identifying potential shortcomings that are not addressed with scaling of such models, as well as issues around catastrophic forgetting in the way current models are used for domain adaptation.

In doing so, we establish state of the art on 5 multilingual downstream tasks (XNLI, PAWS-X, TYDIQA, XQuAD and MLQA) within a model size band, and achieve competitive performance \textit{across} size bands, thereby showing for the first time (to the best of our knowledge) an interesting notion of parameter efficiency: \our{}\textsubscript{XL} outperforms XLM-R\textsubscript{XXL} \citep{xlm-roberta-xlarge} and performs competitively with mT5\textsubscript{XXL} \citep{mT5}, whilst being 5x and 6x smaller respectively (Figure \ref{fig:size_comparison_on_xnli}). Furthermore, our proposed model reduces the gap for English specific tasks: \our{}\textsubscript{XL} achieves 99.3\% GLUE performance and 98.5\% SQuAD 2.0 performance compared to a SoTA English only model in the same size band.

%% file: 1.related_work.tex
\section{Related Work}
Large scale self-supervised learning has emerged as a prominent way of building cross-lingual language models that can be adapted for numerous multilingual downstream applications. Especially for building multilingual encoder transformer \citep{vaswani2017attention} models, two popular paradigms have been Masked language modeling (MLM; \citet{bert, xlm-roberta}) and pre-training encoders as discriminators (ELECTRA; \citet{electraclark2020, xlm-e}), with the latter showing considerable compute efficiency. These approaches can further be improved by leveraging parallel corpora in different ways: \citet{conneau2019cross} propose a Translation Language Modeling task (TLM) wherein the model predicts masked tokens in concatenated translation pairs, \citet{xlm-e} propose a Translation Replaced Token Detection (TRTD) task, an analogous task for Electra-style models. Other approaches include using bitexts to construct code-switched sequences as inputs during pre-training (ALM; \citet{alm}) and for contrastive learning (InfoXLM; \citet{infoxlm}), or using token-level alignments in parallel data to improve cross-lingual modeling \citep[\textit{inter alia}]{amber, chi-etal-2021-improving}. However, all the aforementioned works rely on English-centric bitexts.

\citet{m2m100} show that moving beyond \textit{EN-X} bitexts for Massively Multilingual Machine Translation affords substantial improvements over approaches that rely solely on English-centric data \citep{aharoni-etal-2019-massively,zhang-etal-2020-improving}. The primary factor responsible for this improvement has been the curation of \textit{X-Y} aligned bitext data, constructed by mining bitexts from publicly available web data \citep{wikimatrix,multiccaligned,ccmatrix}. The dataset construction either follows a local mining approach (first aligning documents using heuristics, and then mining parallel bitexts from the aligned documents; used in CCAligned \citep{multiccaligned}), or a global mining approach (all bitexts are embedded in a common vector space, and then aligned candidates are found by looking at the normalized nearest neighbors; used in CCMatrix \citep{ccmatrix}). Due to the added supervision of document alignment, the local mining approaches tend to be less noisy; albeit at the cost of diversity. \citet{m2m100} also propose a sampling strategy for leveraging the \textit{X-Y} bitexts, wherein the marginals are constrained to be similar to what is used for \textit{En-X} bitexts, and show their proposed method improves over uniform sampling. However, as we show in (\S \ref{sec:sampling-distribution}), their proposed strategy has the undesirable artefact of inducing extremely sparse solutions, thereby resulting in data wastage. 

%% file: 2.data-scaling.tex
\section{Leveraging Many-to-Many Bitexts}
\subsection{Dataset}
\label{sec:dataset}
Prior representation learning works usually consider English-centric (\textit{EN-X}) bitexts to improve model quality. Thus, given the emergence of mining based approaches for extracting parallel bitexts from large monolingual datasets that are approximate translations of each other and are multi-way aligned (the source and target languages are not restricted to be English only), in this work we explore leveraging these many-to-many (\textit{X-Y}) bitext datasets for better representation learning. We consider two such publicly available datasets: CCMatrix and multiCCAligned.

\subsection{Sampling Distribution}
\label{sec:sampling-distribution}
\begin{figure*}[!htb]
  \centering

  \subfloat[M2M 100 Sampling]{
  \includegraphics[width=0.48\textwidth, height=0.75\columnwidth]{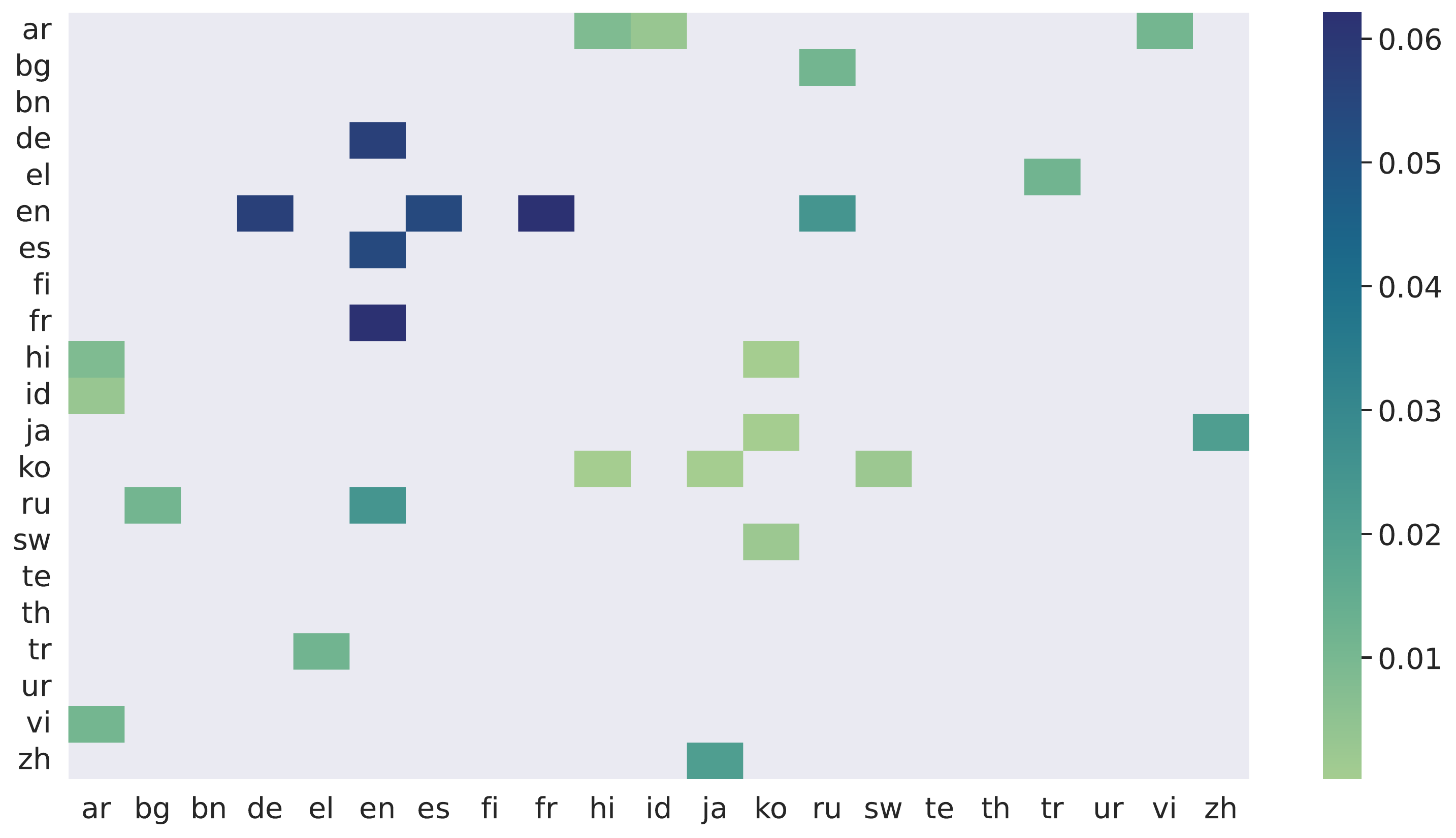}\label{fig:m2m100-density}}
  \subfloat[Proposed Sampling]{
  \includegraphics[width=0.48\textwidth, height=0.75\columnwidth]{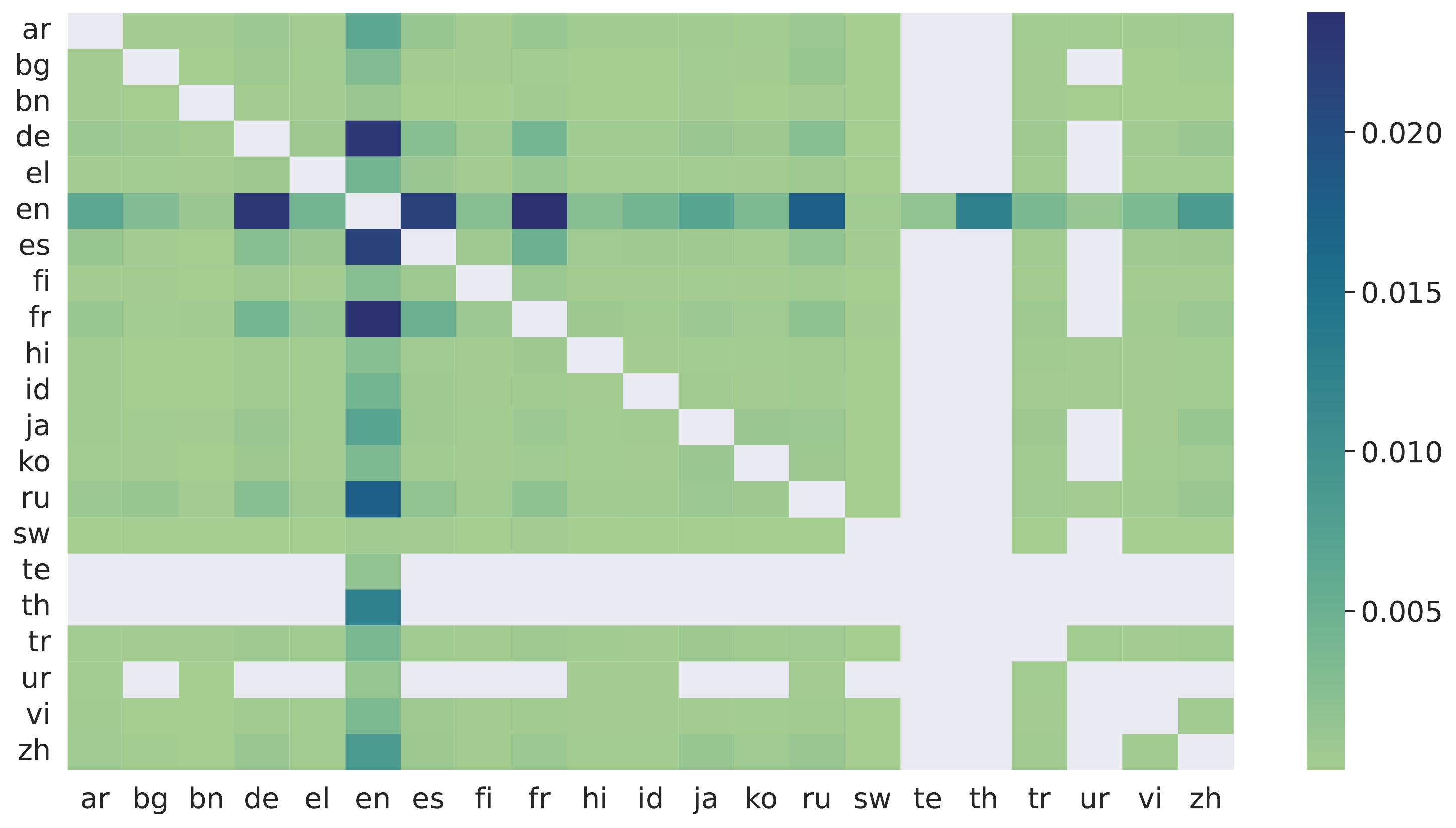}\label{fig:proposed-density}}
  
  \caption{Density plots for our probability distributions for sampling strategies for M2M 100 and our proposed sampling strategy for the 21 languages considered in downstream tasks\label{fig:sampling-density}. For similar plot for all the languages, see Figure \ref{fig:our-density-all} in the Appendix}
\end{figure*}

\label{method:sampling}
A common method used for balancing training data for the \textit{EN-X} framework is using a temperature based exponential sampling approach \citep{aharoni-etal-2019-massively}, wherein the probability of sampling a language is chosen from a temperature smoothed distribution to downsample high resource languages, whilst upsampling low resource languages. This work was extended by \citet{m2m100}, wherein the authors propose Sinkhorn Temperature sampling: given a joint probability matrix $\mathbb{Q}$ across $L\times L$ language pairs ($L$ being the number of unique languages), and the marginal distribution $\mathbf{p}$ of the $L$ languages,  the authors estimate a sampling distribution $\mathbb{P}^{*}$ as:
\begin{equation}
    \max_{\mathbf{P}} \Tr(\mathbf{P}\mathbb{Q}) \mid \mathbf{P}1_{L} = \mathbf{p}^{\frac{1}{T}} = \mathbf{P}^{\top}1_{L} 
    \label{eqn:m2m100}
\end{equation}
where $\Tr$ is the trace operator.
The primary advantage of using this is that $\mathbb{P}^{*}$ can be efficiently estimated with the Sinkhorn-Knopp algorithm and also allows us to set the marginal to be the temperature sampled based distribution which we know works well in practice. The authors found this to work better than uniform sampling.

However, in practice, we observed this to generate extremely sparse sampling distributions: Figure \ref{fig:m2m100-density} show the sparsity induced by the naive application of Eq. \ref{eqn:m2m100}.

We note that one potential way of overcoming the above issue is by modifying the optimization problem to also maximize the entropy of $\mathbf{P}$. Consequently, we propose the following modified optimization objective :

\begin{equation}
\begin{aligned}
    \mathbb{P}^{*} &= \texttt{argmin}_{\mathbf{P}} \Tr\left(P\left(-\log \mathbb{Q}\right)\right) - \mathcal{H}\left(P\right) \\
    & \mid \mathbf{P}1_{L} = \mathbf{p}^{\frac{1}{T}} = \mathbf{P}^{\top} \mathbb{Q}) \mid \mathbf{P}1_{L} = \mathbf{p}^{\frac{1}{T}} = \mathbf{P}^{\top}1_{L}
\label{eqn:kl} 
\end{aligned}
\end{equation}

where $\mathcal{H}(P)$ denotes the entropy of $P$ and $KL(P||Q)$ denotes the Kullback-Leibler divergence between $P$ and $Q$.

This can be solved by using the Sinkhorn-Knopp algorithm for the entropic regularized optimal transport problem \citep{distances2013lightspeed}, by setting the cost matrix to be $-\log(\mathbb{Q} + \epsilon)$ (in practice, since $\mathbb{Q}$ can have zero entries, $\epsilon$ is used for smoothing). Since the cost of assigning a non-zero probability value to a zero entry is extremely high ($-\log\left(\epsilon\right)$), we never observe any entry of  $\mathbb{P}^{*}$ to be non-zero if it's corresponding entry in $\mathbb{Q}$ was zero. In addition, since Eq. \ref{eqn:kl} also maximizes the entropy of $\mathbf{P}$, it encourages its entries to be non-sparse, thereby avoiding the problem present in the solution of Eq. \ref{eqn:m2m100}. In practice, we did not see this losing out on any data: if $\mathbb{Q}$ was non-zero, then $\mathbb{P}^{*}$ was also non-zero (Figure \ref{fig:proposed-density}).

\subsection{Vocabulary Construction}
\label{sec:vocab}
We construct our vocabulary using Sentence Piece Models (SPM) \cite{kudo-richardson-2018-sentencepiece} which cater to language specific complexities (tokenization, accent removal, etc. ). We increase the vocabulary size to 500k tokens to better serve the varied scripts encountered while working in the multilingual setting. For this construction, we follow the VoCAP algorithm \citep{vocap} to quantify the vocabulary capacity for each language separately and account for varied corpora sizes across languages. Better capacity allocation leads to smaller representative sequences (especially for mid and low resource languages) which in-turn improves the computational efficiency of the model. Increasing the size of the vocabulary, however, comes at the cost of inflating the model parameters which is particularly observed in the case of \our{}\textsubscript{Base} and \our{}\textsubscript{Large} where the embeddings layer constitute \textit{80.5\%} and \textit{62.9\%} of the total parameters respectively.

%% file: 3.model-scaling.tex
\section{Pretraining Details}
\label{sec:pretraining-details}
We follow the XLM-E \citep{xlm-e} pretraining approach and only introduce a few architectural changes to improve the overall performance of the model. We use the Transformer model \citep{vaswani2017attention} trained with ELECTRA \citep{electraclark2020} style of replace token detection (RTD) on both monolingual (MRTD) and bitext (TRTD) data. In the current setup of training, we use two Transformer encoders in conjunction:  a generator $G$ and a discriminator $D$, where the generator $G$ is trained with masked language modeling objective (MLM; \citet{bert}) and the discriminator is trained on replaced token detection objective (RTD; \citet{electraclark2020} on all the tokens passing through the generator.

In addition to using the Gated Relative Position Bias introduced in \citet{xlm-e}, we do not mask the [CLS] token and flip bitext language order with probability $p = 0.5$ for the TRTD task.

%% file: 4.experiment-details.tex
\section{Experiments}

\paragraph{Baselines:} We compare the cross-lingual performance of our proposed model against 3 popular cross-lingual models: XLM-R, mT5 and XLM-E (across all model size variations). Note that \citet{xlm-e} use a 250k vocabulary size for XLM-E\textsubscript{Base} and 500k vocabulary for their large and XL variants. As a follow-up, we re-train XLM-E\textsubscript{Base} with the same vocabulary as used by \our{} for a fair comparison. Thus all references to XLM-E\textsubscript{Base} refer to the re-trained model variant with a 500k vocabulary size.\footnote{We also ablate out the impact of the vocabulary change, with Table \ref{tab:ablations} showing that this yields a 1.5 pt gain on XNLI.} For our downstream English evaluation (\S \ref{result:en-centric}), we compare against the SoTA English model METRO-LM\citep{bajaj2022metro}. Note that \citet{bajaj2022metro} also train the models in an ELECTRA style framework, thereby allowing for a fair comparison.

\paragraph{Pretraining Data:} For our monolingual data, we follow \citet{xlm-e} and use the CC-100 dataset\footnote{\url{http://data.statmt.org/cc-100/}} \cite{xlm-roberta,wenzek-etal-2020-ccnet} which contains texts in 100 languages collected from Common Crawl. As mentioned in (\S \ref{sec:dataset}), we explore the utility of the CCMatrix and the multiCCAligned \textit{X-Y} aligned bitext data. CCMatrix consists of 1015 language pairs (97 unique languages) \footnote{For some language pairs that are present in CCAligned and not in CCMatrix, we combine the data from those languages. Since de-duplication is expensive, we don't merge language pairs common to both datasets.}; while the multiCCAligned dataset consists of 2959 language pairs (92 unique languages) \footnote{We filter out languages with less than 50k pairs}. We contrast this against only using \textit{EN-X} bitexts (CCAligned, \citet{multiccaligned}).

\paragraph{Model Size Bands:} While our \textit{base} and \textit{large} models have more parameters when compared with XLM-R, most of the additional parameters come from the increased vocabulary size (\S \ref{sec:vocab}). Concretely, our \textit{base} model has 12 layers and 768 hidden states, while the \textit{large} model has 24 layer and 1024 hidden states, which is identical to XLM-R\textsubscript{Base} and XLM-R\textsubscript{Large} respectively. However, even with the increased parameter count, the computation cost on a text classification task is roughly the same within a model size family (since mapping tokens to an embedding is a lookup operation). Finally, it is noteworthy that even with the increased vocabulary size, the number of parameters for \our{}\textsubscript{XL} is less compared to the XL and XXL variants of both XLM-R and mT5.

\paragraph{Pretraining Setup:} For the base model, we train for 125k steps with a batch size of 8192 for MRTD task and for the large model, we train the model for 500k steps with a batch size of 2048. Finally for the XL model, we train for 150k steps with a batch size of 8192. We use a dynamic batch size for TRTD task which is based on original length of the translated bi-text pair. Please refer Appendix \ref{appendix:pretraining_model_hp} for additional details.  We adopt the standard practice of using a linear warmup schedule for the learning rate and use the Adam \cite{kingma2014adam} optimizer for all the models. Following \citet{meng2021coco}, we do not apply any dropout to the generator.

\paragraph{Cross-lingual Downstream Evaluation:} For evaluating the cross-lingual understanding of the model, we consider 5 multilingual evaluation benchmarks. We consider 2 classification tasks and 3 question answering tasks. For classification, we evaluate on the cross-lingual Natural Language Inference dataset (XNLI; \citet{conneau-etal-2018-xnli}) and the cross-lingual paraphrase adversaries from word scrambling dataset (PAWS-X; \citet{yang-etal-2019-paws}). For cross-lingual question answering, we consider MLQA \citep{lewis2019mlqa}, XQuAD \citep{Artetxe:etal:2019} and TyDiQA-GoldP \citep{clark-etal-2020-tydi}. For all the aforementioned tasks, we perform the evaluation in zero-shot setting, i.e. only using the English data for fine-tuning. To further assess the model's performance when translated data is available, we evaluate the model on the translate-train setup for the classification tasks. 

\paragraph{English Downstream Evaluation:} To further assess \our{}'s performance on English and see how the curse of multilinguality impacts the model, we also assess the model's performance on the commonly used GLUE benchmark \citep{wang-etal-2018-glue}, comprising of 8 tasks: MNLI \citep{williams2017broad}, SST-2 \citep{socher2013recursive}, QNLI \citep{rajpurkar2018know}, MRPC \cite{dolan-brockett-2005-automatically}, CoLA \citep{warstadt2018neural}, QQP , STS-B \cite{cer2017semeval} and RTE. Additionally, we also evaluate the English performance of our model on a question answering task, using the SQuAD 2.0 dataset \citep{squad2}.

Please refer to Appendix \ref{sec:appendix_downstream_perf} for additional details on the datasets.

%% file: 5.results.tex
\section{Results and Analysis}
\label{result}

\begin{table*}[!htb]
    \footnotesize
    \centering
    
    \aboverulesep=0ex
    \belowrulesep=0ex
    \renewcommand{\arraystretch}{1.2}
    \begin{tabular}{@{}l|ccc|cc||cr@{}}
    \toprule
    \multirow{3}{*}{ \makecell[c]{ {\textbf{Model}} } } & \multicolumn{5}{c || }{ \makecell[c]{{\bf Zero-Shot}} } & \multicolumn{2}{c@{} }{ \makecell[c]{{\bf Translate-Train }} } \\
    \cmidrule{2-6} \cmidrule{7-8}
     & \multicolumn{3}{c | } { \makecell[c]{{\bf Question}\\{\bf Answering}} } & \multicolumn{2}{c || }{ \makecell[c]{ {\bf Sentence}\\{\bf Classification} } } & \multicolumn{2}{c@{} }{ \makecell[c]{ {\bf Sentence}\\{\bf Classification} } }  \\
    \cmidrule{2-4} \cmidrule{5-6} \cmidrule{7-8}
    & { {XQuAD} } & { {MLQA} } & { {TyDiQA} } & { {XNLI} } & { {PAWSX} } & { {XNLI} } & { {PAWSX} } \\
    \midrule
    { Metrics } & { {F1/EM} } & { {F1/EM} } & { {F1/EM} }  & { {Acc.} } & { {Acc.} } & { {Acc.} } & { {Acc.} } \\
    \midrule
    { XLM-R\textsubscript{Base} } & { - } & { - } & { - }  & { 76.2 } & { - } & { 79.1 } & { - } \\
    { mT5\textsubscript{Base} } & { 67.0 / 49.0 } & { 64.4 / 45.0 } & { 58.1 / 42.8 } & { 75.4 } & { 86.4 } & { 75.9 } & { 89.3 } \\
    { \baseline{}\textsubscript{Base} } & { 74.3 / 59.2 } & { 68.7 / 50.5 } & { 62.7 / 46.2 } & { 78.1 } & { 87.0 } & { 81.7 } & { 91.1 } \\
    { \our{}\textsubscript{Base} } & {\bf 76.8 / 62.1 } & {\bf 71.3 / 53.2 } & {\bf 67.1 / 51.5 } & {\bf 80.5 } & {\bf 89.7 } & {\bf 84.9 } & {\bf 92.4 }  \\
    \midrule
    { XLM-R\textsubscript{Large} } & { 76.6 / 60.8 } & {65.1 / 45.0} & {71.6 / 53.2}  & { 80.9 } & { 86.4 } & { 83.6 } & { - } \\
    { mT5\textsubscript{Large} } & { 77.8 / 61.5 } & { 71.2 / 51.7 } & { 57.8 / 41.2 } & { 81.1 } & { 88.9 } & { 81.8 } & { 91.2 } \\
    { \baseline{}\textsubscript{Large} } & { 78.7 / 63.1 } & { 72.8 / 54.4 } & { 71.8 / 54.7 } & { 81.3 } & { 89.0 } & { 84.1 } & { 91.9 } \\
    { \our{}\textsubscript{Large} } & { {\bf 79.7 / 64.9 } } & {\bf 74.3 / 55.7 } & {\bf 74.0 / 57.5 } & {\bf 83.0 } & {\bf 90.4 } & {\bf 84.9 } & {\bf 92.4 } \\
    \midrule
    { XLM-R\textsubscript{XL} } & { 80.0 / 64.9 } & { 73.4 / 55.3 } & - & { 82.3 } & { - } & { 85.4 } & { - }\\
    { mT5\textsubscript{XL} } & { 79.5 / 63.6 } & { 73.5 / 54.4 } & { 77.4 / 61.5 }  & { 82.9 } & { 89.6 } & { 84.8 } & { 91.0 } \\
    { \baseline{}\textsubscript{XL} } & { 80.4 / 66.0 } & { 74.3 /  55.8 } & { 76.7 / 60.6 }  & { 83.7 } & { 90.3 } & { 85.5 } & { 92.2 } \\
    { \our{}\textsubscript{XL} } & {\bf 81.3 / 66.3 } & {\bf 75.4 / 56.7 } & {\bf 78.0 / 62.1 } & {\bf 84.8 } & {\bf 91.0 } & {\bf 87.1 } & {\bf 92.6 } \\
    \midrule
    { XLM-R\textsubscript{XXL} } & { 81.1 / 66.3 } & { 74.8 / 56.6 } & { - }  & { 83.1 } & { - } & { 86.0 } & { - } \\
    { mT5\textsubscript{XXL} } & {\bf 82.5 / 66.8 } & {\bf 76.0 / 57.4 } & {\bf 81.0 / 65.6 } & {\bf 85.0 } & {\bf 90.0 } & {\bf 87.8 } & {\bf 91.5 } \\
    \bottomrule
    \end{tabular}
    \caption{Results on sentence-pair classification and question answering tasks. XLM-R metrics and mT5 metrics are reported from \citet{xlm-roberta-xlarge} and \citet{mT5} respectively. Metrics for \baseline{} and \our{} are reported based on median across five fine-tuning runs. 
    Scores of best performing models within a model size band have been highlighted for each task. Full results for all the languages across all tasks can be referred in Appendix \ref{sec:appendix_downstream_perf}. \label{tab:zero-shot}}

\end{table*}

\subsection{Main Results}
\label{sec:main}
Table \ref{tab:zero-shot} presents our proposed model's performance across different model sizes for zero-shot transfer on sentence classification as well as question answering tasks (detailed results for all languages and all tasks can be found in Appendix \ref{app:all-results}). We see that \our{} outperforms the baselines of XLM-E, XLM-R and mT5 across all model sizes, establishing (to the best of our knowledge) the state-of-the-art (SoTA) for all the 5 considered multilingual datasets within the model size bands: with \our{}\textsubscript{Base} outperforming \baseline{}\textsubscript{Base} by 3.1 pts, \our{}\textsubscript{Large} outperforming \baseline{}\textsubscript{Large} by 1.8 pts and \our{}\textsubscript{XL} outperforming \baseline{}\textsubscript{XL} by 0.9 pts (averaged across all 5 datasets). Another interesting observation is that \our{} is competitive across model size families: the \our{}\textsubscript{Base} model out-performs XLM-R\textsubscript{Large} and mT5\textsubscript{Large} variants on 4 out of 5 datasets, similarly the \our{}\textsubscript{Large} outperforms the mT5\textsubscript{XL} model on 4 out of 5 datasets. Furthermore, the \our{}\textsubscript{XL} model outperforms XLM-R\textsubscript{XXL} and is competitive with mT5\textsubscript{XXL} while being 5x and 6x smaller respectively. A practical implication of these better performing smaller models is their easy usage in downstream tasks.

This behaviour is also consistent in the  \textit{Translate-Train} setting where the translated version of the training data is present across all languages for training. Table \ref{tab:zero-shot} presents \our{}'s performance on this setup for sentence classification tasks. We see that even in this setting, \our{} outperforms other models with the same size band, and is competitive across model size bands.

\begin{table}[!htb]
    \footnotesize
    \centering
    
    \aboverulesep=0ex
    \belowrulesep=0ex
    \renewcommand{\arraystretch}{1.2}
    \begin{tabular}{@{}l|c|c@{}}
    \toprule
    { \textbf{Parameter} } & { \textbf{Choice}} & { \textbf{XNLI (Avg)}} \\
    \midrule
     \multirow{2}{*}{ \textbf{Vocabulary Size}} & {250K} & {76.6} \\
     & {500K} & {78.1} \\
     \midrule
    \midrule
    \multirow{3}{*}{ \textbf{Bitext Data}} & {CCAligned} & {78.1} \\
    & {multiCCAligned} & {79.5} \\
    & {CCMatrix} & {80.5} \\
    \midrule\midrule
    \multirow{2}{*}{ \textbf{Training Objective}} & {Masked LM} & {78.4} \\
     & {ELECTRA} & {80.5} \\
    \bottomrule
    \end{tabular}
    \caption{Ablation studies for \our{}. We study the effects of changing few parameters in the pre-training setup while keeping others same. \label{tab:ablations}}
\end{table}

\subsection{Ablations}
\label{sec:ablation}
\paragraph{Different Many-to-Many Datasets}
Table \ref{tab:ablations} shows the impact of moving from English-centric bitexts to \textit{X-Y} bitext data. Using multiCCAligned dataset gives a +1.4 pt improvement on average XNLI performance over the baseline which uses only the CCAligned data, thereby showing that the utility of leveraging multi-way bitext data is not limited to CCMatrix dataset. However, we still see an additional improvement of 1.0 pt with usage of CCMatrix data and we hypothesize this gain to more diversity present in it which in-turn helps in improving the multilingual representations.

\paragraph{Different Pretraining Objectives}
While the gains are more substantial with ELECTRA training objective, Table \ref{tab:ablations} shows that the benefits of having a better quality bitext data is not just restricted to the ELECTRA paradigm of training and can also be observed with the Masked LM objective. For the ablation experiment, we train a base model model with the MLM objective for 125k steps with a batch size of 8192. Comparing this with XLM-R\textsubscript{Base}'s setup, which uses only monolingual data with MLM objective and trains for 1.5M steps (\textit{i.e. 12 times longer}), finally achieving an XNLI (Avg) of 76.2, we observe that introduction of \textit{X-Y} data not only brings performance gains but also significantly improves the training efficiency of these models.

\begin{table*}[!thb]
    \footnotesize
    \centering
    
    \aboverulesep=0ex
    \belowrulesep=0ex
    \renewcommand{\arraystretch}{1.2}
    \begin{tabular}{@{}l@{}|@{}c@{}|@{}c@{}|@{}c@{}|@{}c@{}|@{}c@{}|@{}c@{}|@{}c@{}|@{}c@{}|@{}c@{}|@{}c@{}|@{}r@{}}
    \toprule
    \multirow{2}{*}{{ \textbf{Model} }} & \multicolumn{9}{c|}{{ \textbf{GLUE DEV Single Task} }} & \multicolumn{2}{r@{}}{{ \textbf{SQuAD 2.0} }} \\
    \cmidrule{2-10} \cmidrule{11-12}
    & {{ \makecell[c]{\textbf{MNLI-(m/mm)}\\\textbf{(Acc)}} }} & {{ \makecell[c]{\textbf{QQP}-\textbf{(Acc/F1)}} }} & \makecell[c]{ \textbf{QNLI}\\\textbf{(Acc)}}  & \makecell[c]{ \textbf{SST-2}\\\textbf{(Acc)}}  & \makecell[c]{ \textbf{CoLA}\\\textbf{(MCC)}}  & \makecell[c]{ \textbf{RTE}\\\textbf{(Acc)}}  & {{ \makecell[c]{\textbf{MRPC}\\\textbf{(Acc)}} }} & {{ \makecell[c]{{ \textbf{STS-B} }\\{ \textbf{(SCC)} }}} } & {{ \textbf{AVG} }} & {{ \textbf{EM} }} & {{ \textbf{F1} }} \\
    \midrule
    { METRO-LM\textsubscript{Base} }  & { 90.3 / 90.2} & { 92.4/- } & { 94.4 } & { 95.9 } & { 71.8 } & { 88.1 } & { 91.4 } & { 92.0 } & { 89.5 } & { 85.9 } & { 88.5 } \\
    \midrule
    { \baseline{}\textsubscript{Base} }  & { 86.1 / 86.3 } & { 91.5/88.7 } & { 92.8 } & { 94.0 } & { 67.4 } & { 77.8 } & { 90.2 } & { 91.4 } & { 86.4 } & { 82.3 } & { 85.3 } \\
    \midrule
    { \our{}\textsubscript{Base} }  & { 87.3 / 87.6 } & { 91.9/89.2 } & { 93.3 } & { 94.4 } & { 66.6 } & { 85.2 } & { 90.7 } & { 91.7 } & { 87.6 } & { 83.5 } & { 86.3 } \\
    \midrule
    { $\Delta$ } & { 3.0 / 2.6 } & { 1.3/- } & { 1.1 } & { 1.5 } & { 5.2 } & { 2.9 } & { 0.7 } & { 0.7 } & { 1.9 } & { 2.4 } & { 2.2 } \\
    \midrule
    \midrule
    { METRO-LM\textsubscript{Large} }  & { 91.7 / 91.7 } & { 92.9 } & { 95.8 } & { 96.3 } & { 75.2 } & { 93.1 } & { 92.2 } & { 92.8 } & { 91.4 } & { 88.5 } & { 91.1 } \\
    \midrule
    { XLM-R\textsubscript{Large} }  & { 88.9 / 89.0 } & { 92.3 } & { 93.8 } & { 95.0 } & { - } & { - } & { 89.5 } & { 91.2 } & { - } & { - } & { - } \\
    \midrule
    { \baseline{}\textsubscript{Large} }  & { 89.9 / 90.1 } & { 92.9/90.4 } & { 94.5 } & { 96.8 } & { 73.9 } & { 85.7 } & { 92.1 } & { 92.5 } & { 89.8 } & { 85.6 } & { 88.7 } \\
    \midrule
    { \our{}\textsubscript{Large} }  & { 89.7 / 89.9 } & { 92.7/90.3 } & { 94.7 } & { 95.8 } & { 71.1 } & { 88.4 } & { 91.4 } & { 92.6 } & { 89.6 } & { 85.8 } & { 88.7 } \\
    \midrule
    { $\Delta$ } & { 2.0 / 1.8 } & { 0.2/- } & { 1.1 } & { 0.5 } & { 4.1 } & { 4.7 } & { 0.8 } & { 0.2 } & { 1.8 } & { 2.8 } & { 2.4 } \\
    \midrule
    \midrule
    { METRO-LM\textsubscript{XL} }  & { 92.2 / 92.0 }  & { 93.2 }  & { 96.3 }  & { 97.3 }  & { 76.0 }  & { 93.5 }  & { 91.7 }  & { 93.0 }  & { 91.8 }  & { 89.4 }  & { 92.1 } \\
    \midrule
    { XLM-R\textsubscript{XL} }  & { 90.4/- } & { 92.5 } & { 94.9 } & { 96.6 } & { - } & { - } & { 90.4 } & { - } & { - } & { - } & { - } \\
    \midrule
    { \baseline{}\textsubscript{XL} }  & { 91.1 / 91.2 } & { 92.5/89.9 } & { 94.0 } & { 97.2 } & { 74.7 } & { 91.4 } & { 92.1 } & { 93.2 } & { 90.8 } & { 87.8 } & { 90.7 } \\
    \midrule
    { \our{}\textsubscript{XL} } & { 91.2 / 91.1 } & { 93.0/90.7 } & { 95.8 } & { 96.4 } & { 74.9 } & { 92.8 } & { 91.9 } & { 93.2 } & { 91.2 } & { 88.1 } & { 90.9 } \\
    \midrule
    { $\Delta$ } & { 1.0 / 0.9 } & { 0.2/- } & { 0.5 } & { 0.9 } & { 1.1 } & { 0.7 } & { -0.2 } & { -0.2 } & { 0.6 } & { 1.3 } & { 1.2 } \\
    \bottomrule
    \end{tabular}
    \caption{Results for models on GLUE dev set and SQuAD 2.0 dev set. $\Delta$ represents the performance difference between METRO-LM and \our{} which keeps shrinking we scale up.\label{tab:glue}}
\end{table*}

\subsection{On English Performance of Multi-Lingual Models}
\label{result:en-centric}

Given the strong performance of multilingual models on the English subset of XNLI, one interesting question that arises is how does model scaling impact the performance on English centric downstream tasks. In order to evaluate that, we measure the performance of \our{} on the commonly used GLUE benchmark \citep{wang-etal-2018-glue} and the SQuAD 2.0 benchmark. To compare the multilingual model performance on English, we also consider English specific encoder models trained in an Electra pre-training paradigm. Specifically, we consider the Base, Large, XL and XXL models presented in \citep{bajaj2022metro}.

Table \ref{tab:glue} shows the performance of our proposed method against the SoTA monolingual as well as other multilingual baselines. As observed in the results, with  an increase in the number of parameters, we see that the gap in the performance of an English centric model and a multilingual model decreases, with the XL model being just 0.6 points behind on GLUE and 1.3 points on SQuAD 2.0. We hypothesize that an increase in model capacity alleviates the issues caused by the curse of multilinguality \citep{xlm-roberta}; and when that is the case, English performance actually benefits from the presence of other languages in the training data.

It is noteworthy that the even for the English language performance, having an \textit{X-Y} centric data is more beneficial compared to an \textit{EN-X} data (\baseline{} vs \our{}). Furthermore, our proposed method outperforms XLM-R on large and XL sizes.

\subsection{Performance Across Language Families}
\label{sec:language-families}
\begin{figure}[!htb]
  \centering
  \includegraphics[width=0.48\textwidth,]{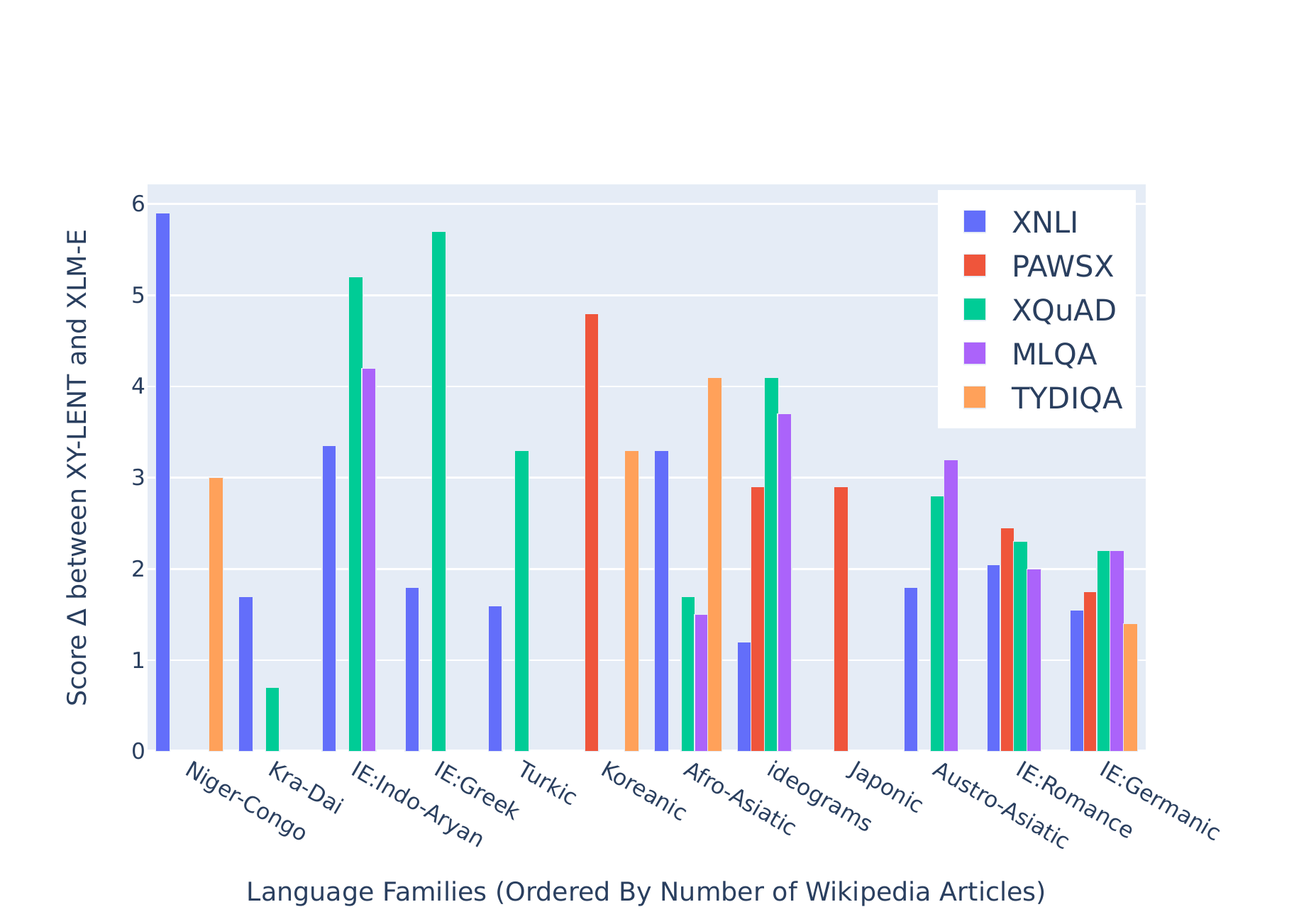}\label{fig:our-density}
  
  \caption{Performance $\Delta$ between \baseline{} and \our{} across language families\label{fig:perf-delta}}
\end{figure}

Figure \ref{fig:perf-delta} shows the performance the delta of performance between \baseline{} and \our{} across different language families. Following \citet{hu2020xtreme}, we use the number of Wikipedia articles as a proxy for a language family being high or low resource. As can be seen, leveraging X-Y bitexts helps improves performance consistently across language families.

\subsection{Crosslingual Transfer Gap}
\label{sec:crosslingual-transfer-gap}
\begin{table}[!htb]
    \footnotesize
    \centering
    
    \aboverulesep=0ex
    \belowrulesep=0ex
    \renewcommand{\arraystretch}{1.2}
    \begin{tabular}{@{}l@{}c@{}c@{}c@{}c@{}r@{}}
    \toprule
         {{\bf Model } } & { {\bf XQuAD }} & { {\bf MLQA }} & { {\bf TyDiQA }} & { {\bf XNLI } } & { {\bf PAWS-X}}  \\
         \midrule
         {MBERT } & { 25.0 } & { 27.5 } & { 22.2 } & { 16.5 } & { 14.1} \\
         \midrule
         {XLM-R } & { 15.9 } & { 20.3 } & { 15.2 } & { 10.4 } & { 11.4} \\
         \midrule
         {XLM-E } & { {\bf 14.9} } & { {\bf 19.2} } & { 13.1 } & { 11.2 } & { 8.8} \\
         \midrule
         {\our{} } & { 15.3 } & { 19.9 } & { {\bf 8.6} } & { {\bf 7.8} } & { {\bf 6.8}} \\
    \bottomrule
    \end{tabular}
    \caption{Crosslingual Transfer Gap scores on 5 multilingual benchmark tasks. A lower score indicates better cross-lingual transfer. For QnA datasets, this is computed using the EM scores. \label{tab:crosslingual-transfer}}
    \label{tab:my_label}
\end{table}
In order to further evaluate the cross-lingual transferrability of our model, we follow \citet{hu2020xtreme} and evaluate the cross-lingual transfer gap (the difference between the performance on the English test set and the average test set performance for other languages) for \our{}\textsubscript{Base}. This score indicates how much end task knowledge is not transferred to other languages post fine-tuning, with a smaller gap  indicating better transferrability. As seen in Table \ref{tab:crosslingual-transfer}, \our{} achieves lower scores on 3 out of 5 tasks, thereby demonstrating strong transferrability.

\subsection{Using Training Dynamics to Explore Dataset Quality}
\label{result:dataset}
So far we have seen that leveraging \textit{X-Y} aligned bitexts improves model quality. In this section, we consider the inverse direction: whether training dynamics of representation learning models can be used to identify dataset artifacts. Given these bitext datasets span over 1000 language pairs, a manual inspection of these datasets is extremely hard. Thus an automated method for spot-checking the dataset quality is quite valuable.

To do so, we first train a model in line with the methodology presented by \citet{zhou-etal-2021-distributionally} for Distributionally Robust Multilingual Machine Translation. Specifically, we train \our{} with the following modified objective:

\begin{equation}
\begin{aligned}
    \min_{\theta_{D}, \theta_{G}} \sup_{\mathbf{P}: \chi^{2}\left(\mathbf{P}, \mathbb{Q}\right)\leq \rho} \sum_{i} p_i (\mathcal{L}_{D}(\mathbf{x}; \theta_{D}) + \\
    \lambda \mathcal{L}_{G}(\mathbf{x}; \theta_{G}))
\end{aligned}
\end{equation}

Here $\mathcal{L}_{G}$ and $\mathcal{L}_{D}$ refer to the generator and discriminator losses respectively (\S \ref{sec:pretraining-details}), $\mathbf{P}$ is the joint distribution over the bitext language pairs that we want to estimate (i.e $\mathbb{P} = {p_i \mid 1\leq i \leq L\times L; \sum_{i} p_i = 1}$); and $\mathbb{Q}$ is the original training distribution (i.e the probability distribution over the bitexts when the training starts, equal to $\mathbb{P}^{*}$ as estimated in \S \ref{sec:sampling-distribution}). At a high level, the objective minimizes the training loss over a $\chi^{2}$ ball around the original training distribution, with the supremum up-weighing language pairs with higher loss values, and down-weighing languages with lower loss values \footnote{Table \ref{tab:xnli-main-all} in the Appendix shows that such an approach achieves reasonable performance on XNLI.}. We train a model with the Distributional Robustness Optimization objective (DRO) using Iterated Best Response strategy, as proposed by \citet{zhou-etal-2021-distributionally} and resample 10 times throughout the training. We hypothesize that the two extremities (i.e language pairs that are highly upsampled as well as those that are downsampled) would be bitext datasets of interest for spot-checking.  

\begin{figure}[!htb]
    \includegraphics[scale=0.45]{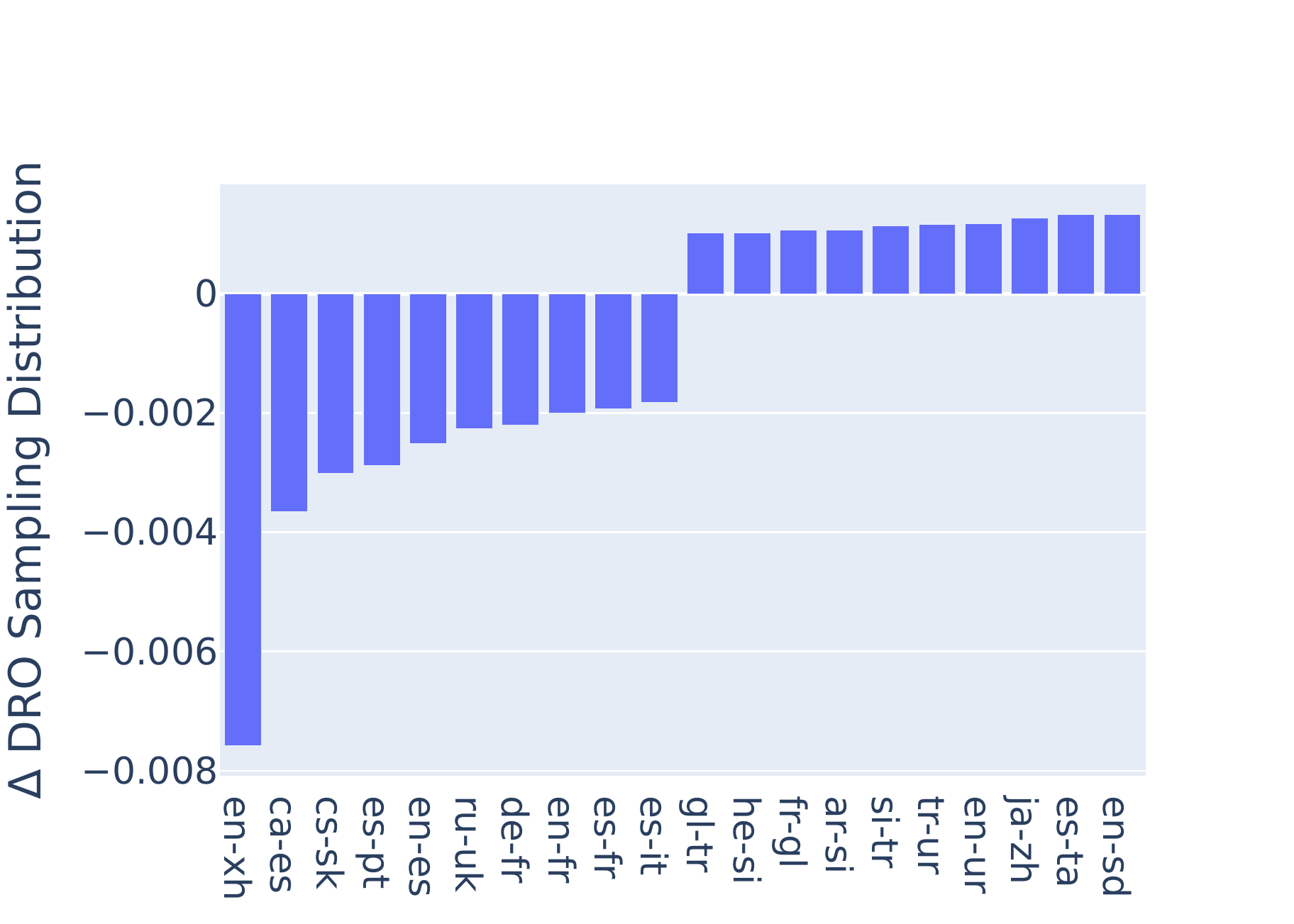}
    \caption{ The 10 upsampled and downsampled languages when the model is trained with the DRO objective\label{fig:dro}}
\end{figure}

Figure \ref{fig:dro} presents the top 10 upsampled and 10 downsampled languages between the initial and final language distributions. Manual inspection of these language pairs shows that our hypothesis indeed holds true: we observe that the  translations for English and Xhosa (en - xh) are extremely noisy and aligned with non-sensical text, with multiple different English sentences being aligned to the same Xhosa sentence\footnote{This can potentially be a manifestation of the well studied Hubness issue for Nearest Neighbor lookups in high dimensional spaces \citep{radovanovic2010hubs, Dinu-hubness})}. Bitexts for Catalan and Spanish (ca - es) and Czech and Slovak (cs - sk) are near duplicates, since the language pairs are very similar. Both of these issues can cause the TRD task to be trivial, explaining the downsampling. Similarly, looking at languages that are up-sampled, we observe a lot of translation quality noise in bitexts for Spanish and Tamil (es - ta), Turkish and Urdu (tr - ur) and Sinhala and Turkish (si - tr).

%% file: 6.conclusion.tex
\section{Conclusion}
In this work, we introduced a family of models which achieve SoTA performance over 5 multilingual benchmarks compared to other models belonging to similar model size bands and are competitive across the bands. Our \our{}\textsubscript{XL} model outperforms XLM-R\textsubscript{XXL} and is competitive with mT5\textsubscript{XXL} being 5x and 6x smaller respectively. Furthermore, the XL model variant also achieves 99.3\% and 98.5\% of the current best performant models on GLUE and SQuAD 2.0 respectively, thereby aiding in reducing the curse of multilinguality. The performance gains are consistent across language families.

%% file: 7.limitations.tex
\section{Limitations}

\begin{table*}[!bht]
    \footnotesize
    \centering
    \aboverulesep=0ex
    \belowrulesep=0ex
    \renewcommand{\arraystretch}{1.2}
    \begin{tabular}{@{}l@{}|@{}c@{}|@{}c@{}|@{}c@{}|@{}c@{}|@{}c@{}|@{}c@{}|@{}c@{}|@{}c@{}|@{}c@{}|@{}c@{}|@{}c@{}|@{}c@{}|@{}r@{}}
    \toprule
    { \textbf{Model} } & { \textbf{Avg} } & { \makecell[c]{\textbf{Avg w/o}\\\textit{\bf en}} } & { \textbf{en} } & { \textbf{aym}  } & { \textbf{bzd}  } & { \textbf{cni}  } & { \textbf{gn}  } & { \textbf{hch}  } & { \textbf{nah}  } & { \textbf{oto}  } & { \textbf{quy}  } & { \textbf{shp}  } & { \textbf{tar} } \\
    \midrule
    { XLM-R\textsubscript{Base} } & { 39.4 } & { 38.5 } & { 85.8 } & { 36.1 } & { 39.7 } & { 37.9 } & { 39.5 } & { 37.2 } & { 42.6 } & { 37.8 } & { 37.2 } & { 40.5 } & { 36.4 } \\
    \midrule
    { \baseline{}\textsubscript{Base} } & { 44.8 } & { 40.6 } & { 87.5 } & { 40 } & { 38.8 } & { 41.7 } & { 43.6 } & { 38 } & { 43.8 } & { 39.8 } & { 41.7 } & { 42.3 } & { 35.9 } \\
    \midrule
    { \our{}\textsubscript{Base}} & { 45.5 } & { 41.6 } & { 84.4 } & { 40.7 } & { 40.7 } & { 42.9 } & { 42.5 } & { 38.9 } & { 45.5 } & { 40.9 } & { 42.1 } & { 43.9 } & { 37.6 } \\
    \midrule
    { \our{}\textsubscript{XL} } & { 47.2 } & { 42.8 } & { 90.8 } & { 42.1 } & { 42.5 } & { 45.6 } & { 42.9 } & { 41.2 } & { 45.0 } & { 41.3 } & { 42.7 } & { 46.9 } & { 37.9 } \\
    \bottomrule
    \end{tabular}
    \caption{Performance of models on the AmericasNLI\label{tab:americas-nli} dataset. Note that model scaling does not seem to improve performance as much for these unseen languages.}
\end{table*}
\label{limitations}

Even though \our{} paves the way towards better general-purpose multilingual representation foundation models, in this section, we highlight the limitations associated with this work. We first expound upon the limitations associated with self-supervised learning on large web extracted corpora. Then we show that while \our{} achieves strong performance on multiple multilingual benchmarks, when the downstream task involves unseen (during pretraining) languages, the performance drops by a substantial margin. Finally, we show the potential limitation associated with a common methodology used for domain adaptation associated with leveraging these multilingual foundation models, illustrating how catastrophic forgetting exacerbates certail issues pertaining to low resource language performance. 

\subsection*{Training Data}
\our{} uses CC-100 which a static multilingual corpus extracted from Common Crawl for 100 languages. As noted by \citet{wenzek-etal-2020-ccnet}, several data filtering strategies have been applied to remove duplicated documents, paragraphs with high ratio of punctuations, digits and profanities, the resultant data may still result in many potential biases requiring further analysis. Additionally, these issues might be aggravated for models that leverage bitext data, since the bitexts themselves are mined from web crawls, and thus potentially have all the associated biases, stereotypes and other associated harms. Furthermore, the raw data was compiled from static Common Crawl snapshots from January, 2020 to December, 2020 and hence may not include information about some of the recent events such as COVID-19.

\subsection*{Performance on Unseen Languages}

Given the performance improvements observed with scaling, we investigate how it impacts extremely low resource languages which are not present in the pre-training data. In order to do so, we consider our model's performance on the AmericasNLI dataset \cite{ebrahimi-etal-2022-americasnli} which extends the XNLI dataset to 10 Indigenous languages of the Americas.

Table \ref{tab:americas-nli} presents the results on the AmericasNLI dataset. As can be seen, \our{} does outperform XLM-R, indicating that better representation learning also benefits these extremely low resource languages. However, we do not see an increase in performance while scaling our models. Specifically, the performance of \our{}\textsubscript{Base} and \our{}\textsubscript{XL} model is nearly the same, and substantially worse that the performance observed on the XNLI dataset. This indicates that, while parameter scaling can help improve performance on languages that the model has seen during pre-training, it does not automatically improve performance in the extremely low-resource regime \footnote{Note that since the tokenizer is a sentencepiece tokenzier, there are extremely few \texttt{UNK} words in the low-resource languages. Consequently, the poor performance is not explained by excessive occurrences of \texttt{UNK} tokens}. Thus, while model scaling allows for improvements across numerous dimensions, it is far from a panacea, especially if not done in conjunction with data scaling efforts. To be able to improve performance for unseen languages, an intervention would need to be made at the data collection efforts during pre-training, which we aim to assess in future works.

\subsection*{Continued Training for Domain Adaptation in Pre-Trained Encoders}

\begin{table*}[!hbt]
    \footnotesize
    \centering
    \aboverulesep=0ex
    \belowrulesep=0ex
    \renewcommand{\arraystretch}{1.2}

    \begin{tabular}{@{}l@{}|@{}c@{}|@{}c@{}|@{}c@{}@{}c@{}@{}c@{}@{}c@{}@{}c@{}@{}c@{}@{}c@{}@{}c@{}@{}c@{}@{}c@{}@{}c@{}@{}c@{}@{}c@{}@{}c@{}@{}c@{}@{}r@{}}
    \toprule
    \textbf{{ Model }} & \textbf{{ CT }} & \textbf{{ Avg }} & \textbf{{ en }} & \textbf{{ fr }} & \textbf{{ es }} & \textbf{{ de }} & \textbf{{ el }} & \textbf{{ bg }} & \textbf{{ ru }} & \textbf{{ tr }} & \textbf{{ ar }} & \textbf{{ vi }} & \textbf{{ th }} & \textbf{{ zh }} & \textbf{{ hi }} & \textbf{{ sw }} & \textbf{{ ur }} \\
    \midrule
    \multirow{5}{*}{ \makecell[c]{\our{}\\MLM + TLM} } & {\ding{55}} & { 78.4 } & { 86.2 } & { 81.5 } & { 82.9 } & { 81.5 } & { 80.6 } & { 81.8 } & { 79.8 } & { 77.4 } & { 77.9 } & { 78.3 } & { 75.2 } & { 78.3 } & { 73.3 } & { 72.8 } & { 68.0 } \\
    \cmidrule{2-18}
    & {\ding{51}} & { 67.5 } & { 85.7 } & { 73.9 } & { 74.5 } & { 71.4 } & { 71.7 } & { 70.9 } & { 71.2 } & { 57.8 } & { 65.1 } & { 66.7 } & { 68.5 } & { 71.8 } &  { 60.6 } & { 45.8 } & { 57.1 } \\
    \cmidrule{2-18}
    & { \makecell[c]{Relative\\$\Delta(\%)$} } & { 13.9 } & { 0.6 } & { 9.3 } & { 10.1 } & { 12.4 } & { 11.0 } & { 13.3 } & { 10.8 } & { 25.3 } & { 16.4 } & { 14.8 } & { 8.9 } & { 8.3 } & { 17.3 } & { 37.1 } & { 16.0 } \\
    \cmidrule{2-18}
    \cmidrule{2-18}
    & { \ding{51} w/ low LR } & { 73.5 } & { 85.9 } & { 78.2 } & { 78.6 } & { 76.2 } & { 76.8 } & { 77 } & { 76 } & { 68 } & { 72.2 } & { 73.7 } & { 73.8 } & { 76.2 } & { 68.7 } & { 57.4 } & { 64.4 } \\
    \cmidrule{2-18}
    & { \makecell[c]{Relative\\$\Delta(\%)$} } & { 6.3 } & { 0.3 } & { 4.0 } & { 5.2 } & { 6.5 } & { 4.7 } & { 5.9 } & { 4.8 } & { 12.1 } & { 7.3 } & { 5.9 } & { 1.9 } & { 2.7 } & { 6.3 } & { 21.2 } & { 5.3 } \\
    \midrule
    \midrule
    \multirow{3}{*}{ \our{}\textsubscript{Base} } & {\ding{55}} & { 80.3 } &  { 87.9 } &  { 83.4 } &  { 84.4 } &  { 82.9 } &  { 82.6 } &  { 83.1 } &  { 81.1 } &  { 79.5 } &  { 79.5 } &  { 80.0 } &  { 77.7 } &  { 80.1 } &  { 76.4 } &  { 75.3 } &  { 71.3 } \\
    \cmidrule{2-18}
    & {\ding{51}} & { 75.6 } & { 87.5 } & { 80.5 } & { 81.6 } & { 78.2 } & { 79.3 } & { 79.4 } & { 76.8 } & { 72.4 } & { 74.2 } & { 76.4 } & { 75.3 } & { 78.7 } & { 70.3 } & { 58.8 } & { 65.1 } \\
    \cmidrule{2-18}
    & { \makecell[c]{Relative\\$\Delta(\%)$} } & { 5.9 } & { 0.5 } & { 3.5 } & { 3.3 } & { 5.7 } & { 4.0 } & { 4.5 } & { 5.3 } & { 8.9 } & { 6.7 } & { 4.5 } & { 3.1 } & { 1.7 } & { 8.0 } & { 21.9 } & { 8.7 } \\
    \bottomrule
    \end{tabular}
    \caption{Drop in cross-lingual zero-shot performance before and after continued training (CT). For MLM, we show with original LR and lower LR. $\Delta$ measured as a relative (\%) drop compared to no CT\label{tab:drop-xlingual-perf}}
\end{table*}

In recent years, continued training on domain specific corpora has been considered a viable approach for domain adaptation of MLM style pre-trained models \citep{gururangan-etal-2020-dont, yao-etal-2021-adapt} where the core idea is to continue train the pre-trained model on domain specific corpora with the goal of improving in-domain downstream evaluation.

We first show that this phenomenon can be extended to models pretrained with an ELECTRA style training objective. Concretely, we apply domain adaptation in the biomedical domain where we continue to train our \our{}\textsubscript{Base} as well as \our{}\textsubscript{MLM + TLM} model on the PubMed data presented in \citet{yao-etal-2021-adapt}, and evaluate it on the ChemProt task (which aims at extracting relations between chemicals and proteins) presented in \citet{gururangan-etal-2020-dont} as the in-domain downstream task.

\begin{table}[!hbt]
    \footnotesize
    \centering
    \aboverulesep=0ex
    \belowrulesep=0ex
    \renewcommand{\arraystretch}{1.2}
    \begin{tabular}{@{}c@{}|@{}c@{}|@{}c@{}}
        \toprule
         { \textbf{Model} } & { \makecell[c]{\textbf{Acc.}\\\textbf{(w/o Contd. Train)}} } &  { \makecell[c]{\textbf{Acc.}\\\textbf{(Contd. Train)}} } \\
         \midrule
         { \makecell[c]{\our{}\\MLM + TLM} } & { 82.0 } & { 86.0 } \\
         \midrule
         { \our{}\textsubscript{Base} } & { 81.6 } & { 86.2 } \\
         \bottomrule
    \end{tabular}
    \caption{Domain Specific Downstream task: Accuracy on Chemprot dataset\label{tab:domain-adapt}}
\end{table}

\begin{figure}[!htb]
    \includegraphics[scale=0.225]{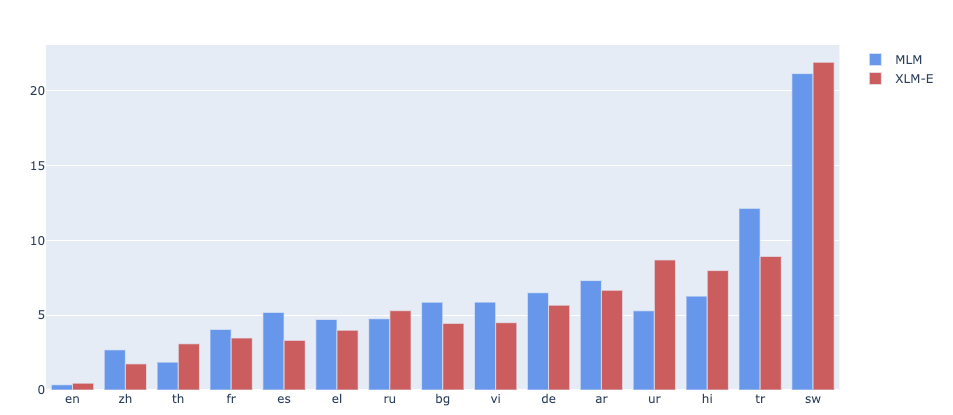}
    \caption{Relative Zero-Shot performance drop with continued training for MLM and ELECTRA style models\label{fig:rel-drop}}
\end{figure}

We observe that the continued training approach presented in \citet{gururangan-etal-2020-dont} for the ELECTRA style models, using the same peak learning rate as used during pre-training, results in divergence. Interestingly, this neither happens for the generator of the ELECTRA model nor for the MLM style pre-trained model. Thus, for an ELECTRA style continued training setup, we posit reducing the peak learning rate to be a crucial change. Table \ref{tab:domain-adapt} shows the performance on the downstream task post the continued training approach and unsurprisingly it helps with improving in-domain performance.

However, given the multilingual nature of such models, we test the multilinguality of these models before and after continued training; using cross-lingual zero-shot XNLI as a proxy for multilingual model quality. Table \ref{tab:drop-xlingual-perf} shows the drop in performance across all languages pre and post continued training. We first note that this drop in performance is present for both MLM and ELECTRA style of models, and thus is not an artifact of the pre-training objective. We observe that the drop in performance is not uniform across all languages and the drop is worse for MLM style models (with using the same peak learning rate suffering more from this issue; Table \ref{tab:domain-adapt}). While we expect the drop in English performance to be relatively less, we do see that the drop is substantially more for the mid and low resource languages (especially Hindi, Turkish, Urdu and Swahili; see Fig. \ref{fig:rel-drop}). While this can potentially be ameliorated by using techniques like Adapters \citep{houlsby2019parameter} etc., we would like to draw attention towards the fact that general purpose continued training does suffer from this issue.

%% file: Appendix.tex
\clearpage

\section*{Appendix}
\section{Pre-Training and Model Hyperparameters}
\begin{table}[!htb]
\centering
\small
\renewcommand\tabcolsep{3.5pt}
\begin{tabular}{lrrr}
\toprule
Hyperparameters & Base & Large & XL \\ \midrule
Layers & 12 & 24 & 48 \\
Hidden size & 768 & 1,024 & 1,536 \\
FFN inner hidden size & 3,072 & 4,096 & 6,144 \\
Attention heads & 12 & 16 & 24 \\
\bottomrule
\end{tabular}
\caption{Model hyperparameters of \our{} discriminators across different sizes.}
\label{table:d-hparam}
\end{table}

\begin{table}[!htb]
\centering
\small
\renewcommand\tabcolsep{3.5pt}
\begin{tabular}{lccr}
\toprule
Hyperparameters & Base & Large & XL \\ \midrule
Training steps & 125K & 500K & 150K \\
Batch tokens per task & 4M & 1M & 4M \\
Adam $\epsilon$ & 1e-6 & 1e-6 & 1e-6\\
Adam $\beta$ & (0.9, 0.98) & (0.9, 0.98) & (0.9, 0.98) \\
Learning rate & 8e-4 & 2e-4 & 1e-4 \\
Learning rate schedule & Linear & Linear & Linear \\
Warmup steps & 10,000 & 10,000 & 10,000 \\
Gradient clipping & 2.0 & 1.0 & 1.0  \\
Weight decay & 0.01 & 0.01 & 0.01 \\
\bottomrule
\end{tabular}
\caption{Hyperparameters used for pre-training \our{}.
}
\label{table:pt-hparam}
\end{table}

\label{appendix:pretraining_model_hp}
Table \ref{table:d-hparam} shows the hyper-parameters of \our{} across various model sizes. All the models are trained with a vocabulary size of 500K and we use batch size of 1M or 4M tokens based on model size as mentioned in Table \ref{table:pt-hparam}. For multilingual replace token detection task we work with a fixed input sequence length of 512 and hence maintains a constant batch size. For translation replace token detection task, the input sequence length is dynamically set as the length of original translation pair and the max one is chosen across the batch. For the base and large models, we train on 128 Nvidia A100-40GB GPU cards, and for the XL model, we use 512 Nvidia A100-80GB GPU cards.

\section{Downstream Performance}
\label{sec:appendix_downstream_perf}
For evaluating cross lingual understanding, we consider five multilingual evaluation benchmarks.
We use XNLI (Cross-Lingual Natural Language Inference) and PAWS-X for classification and XQuAD, MLQA and TyDiQA-GP for question answering. 
Additionally, we use GLUE benchmark and SQuAD2.0 to evaluate the English performance of our model.

\begin{figure*}[!bht]
  \centering

  \subfloat[M2M 100 Sampling]{
  \includegraphics[width=0.50\textwidth]{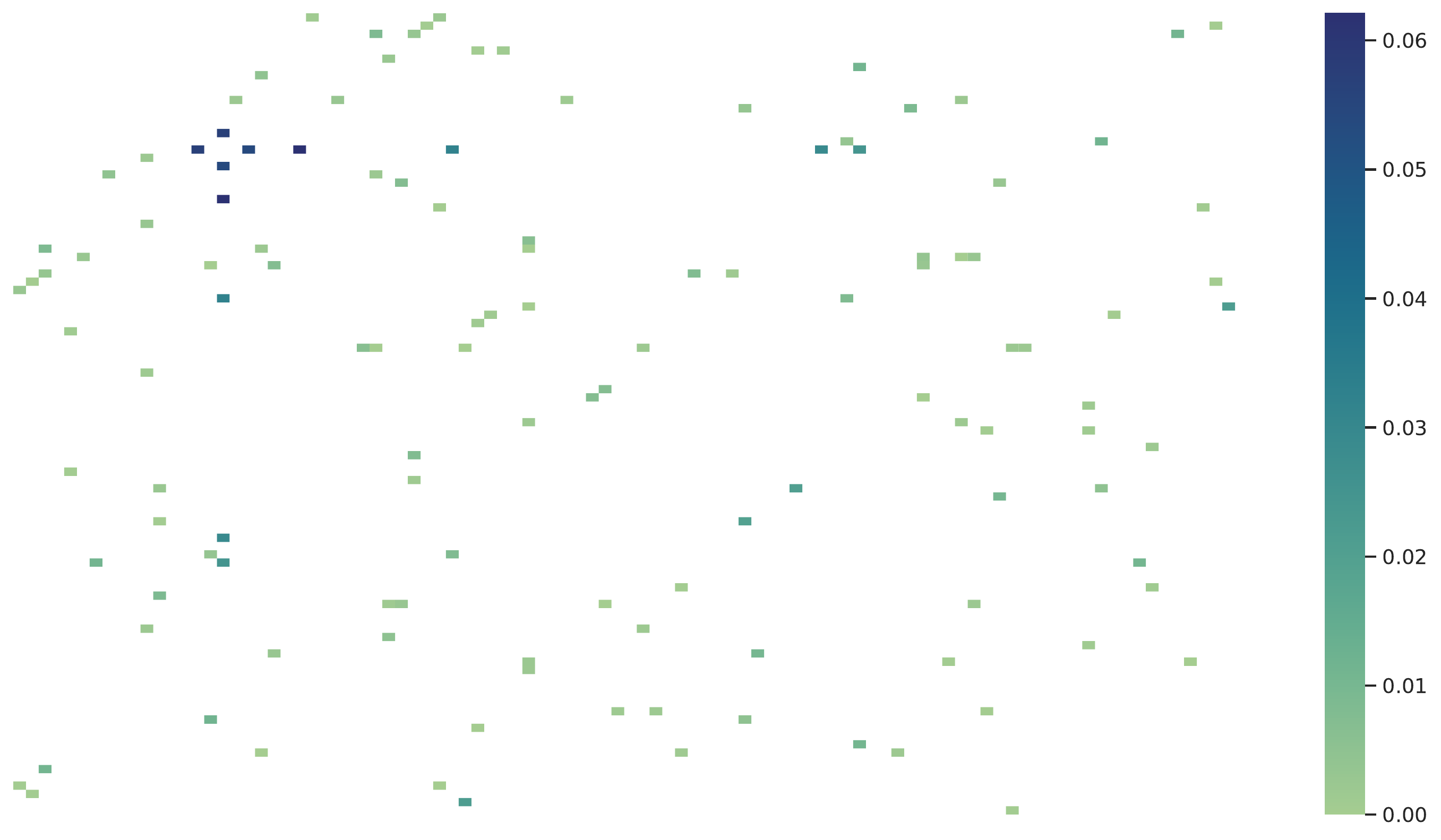}\label{fig:m2m100-density-all}}
  \subfloat[Proposed Sampling]{
  \includegraphics[width=0.50\textwidth]{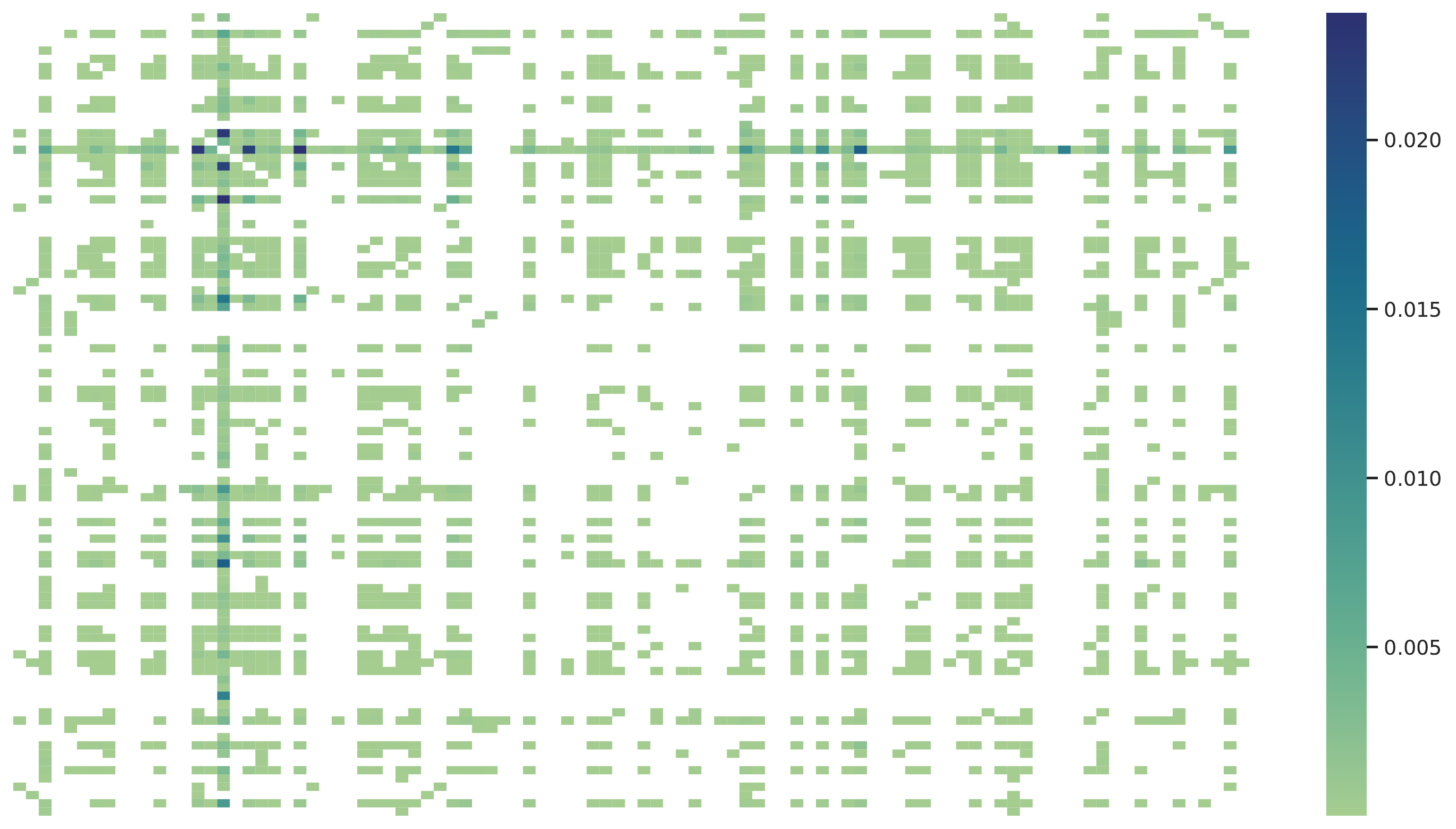}\label{fig:our-density-all}}
  
  \caption{Density plots for our probability distributions obtained for sampling strategies for M2M 100 vs our proposed strategy for all languages in the training set.\label{fig:sampling-density-all}}
\end{figure*}

\paragraph{XNLI}
The XNLI dataset~\cite{conneau-etal-2018-xnli} comes with ground-truth dev and test sets in 15 languages, and a ground-truth English training set. The training set has been machine-translated to the remaining 14 languages, providing synthetic training data for these languages as well. We evaluate our model on cross-lingual transfer from English to other languages in two modes: (i)\textit{zero-shot}: the model is fine-tuned only using the English training data and (ii) \textit{translate-train-all}: the English training set is machine-translated to each language and we fine-tune a multilingual model on all training sets. For translations, we use the original XNLI data for consistency.

\paragraph{PAWS-X}
The PAWS (Paraphrase Adversaries from Word Scrambling) dataset ~\cite{zhang-etal-2019-paws} requires to
determine whether two sentences are paraphrases. We use the subset of the PAWS dev and test sets 
translated to six other languages by professional translators, dubbed as PAWS-X ~\cite{yang-etal-2019-paws}
for evaluation, while using the PAWS set for training.

\paragraph{XQuAD}
The English SQuAD v1.1\cite{rajpurkar-etal-2016-squad}  requires identifying the answer to a
question as a span in the corresponding paragraph. In XQuAD\cite{Artetxe:etal:2019}, a subset
of the English  dev set was translated into ten other languages by professional
translators which is then used for evaluation.

\paragraph{MLQA}
The Multilingual Question Answering\cite{lewis2019mlqa} dataset is another cross-lingual question answering
dataset. In this dataset, the evaluation data for English and six other languages was obtained by automatically mining target language 
sentences that are parallel to sentences in English from Wikipedia, crowd-sourcing annotations in
English, and translating the question and aligning the answer spans in the target languages. We use the SQuAD v1.1\cite{rajpurkar-etal-2016-squad} training data for training and evaluate on the test data of the corresponding task.

\paragraph{TyDiQA-GP}
We use the gold passage version of the Typologically Diverse Question Answering\cite{clark-etal-2020-tydi} dataset, a benchmark for information-seeking question answering, which covers nine languages. The gold passage version is a simplified version of the primary task,
which uses only the gold passage as context and excludes unanswerable questions. It is thus similar to XQuAD and
MLQA, while being more challenging as questions have been written without seeing the answers, leading to 3× and
2× less lexical overlap compared to XQuAD and MLQA respectively. We use the English training data for training
and evaluate on the test sets of the target languages.

\paragraph{GLUE and SQuAD 2.0}
We evaluate English performance of our model on the GLUE benchmark \cite{wang-etal-2018-glue} 
which is a benchmark of eight diverse NLU tasks spanning over single-sentence tasks (CoLA, SST-2), similarity and paraphrase tasks (MRPC, STS-B, QQP) and inference tasks (RTE, MNLI, QNLI). The benchmark is also varied in terms of the training data sizes across tasks which makes it an effective benchmark for testing NLU capabilities of a pretrained model in a robust fashion. We also evaluate the English performance on SQuAD 2.0 \citep{squad2} task which is a collection of 100k crowdsourced question/answer pairs collected from Wikipedia where given a passage and a question, the task is to predict the answer span in the passage. The task also has the possibility that no answer exists, making the problem more grounded.

\section{Sampling Sparsity Across All Language Pairs}
Figure \ref{fig:sampling-density-all} shows the sampling distribution as induced by the M2M sampling method and by our proposed method for all language pair directions. Our proposed method induces a much less sparse distribution, resulting in less data wastage.

\section{Detailed Performance on All Tasks and Languages}
\label{app:all-results}
We present the detailed results associated with all tasks and languages in this section.

\begin{table}[!htb]
    \footnotesize
    \centering
    \aboverulesep=0ex
    \belowrulesep=0ex
    \renewcommand{\arraystretch}{1.2}
    \begin{tabular}{@{}l@{}|@{}c@{}|@{}c@{}@{}c@{}@{}c@{}@{}c@{}@{}c@{}@{}c@{}@{}r@{}}
    \toprule
    { \textbf{Model} } & { \textbf{Avg} } & { \textbf{en} } & { \textbf{de} } & { \textbf{es} } & { \textbf{fr} } & { \textbf{ja} } & { \textbf{ko} } & { \textbf{zh} } \\
    \midrule
    \midrule
    \multicolumn{9}{@{}l}{\textit{Zero-shot Crosslingual Transfer}} \\
    \midrule
    \midrule
    \multicolumn{9}{@{}c@{}}{\textbf{Base} } \\
    \midrule
    { mT5 } & { 86.4 } & { 95.4 } & { 89.4 } & { 89.6 } & { 91.2 } & { 79.8 } & { 78.5 } & { 81.1 } \\
    \midrule
    { \baseline{} } & { 87.0 } & { 94.9 } & { 89.4 } & { 90.3 } & { 90.5 } & { 81.1 } & { 78.9 } & { 83.8 } \\
    \midrule
    { \our{} } & { 89.7 } & { 95.5 } & { 92.3 } & { 92.5 } & { 93.2 } & { 84 } & { 83.7 } & { 86.7 } \\
    \midrule
    \multicolumn{9}{@{}c@{}}{\textbf{Large} } \\
    \midrule
    { mT5 } & { 88.9 } & { 96.1 } & { 91.3 } & { 92 } & { 92.7 } & { 82.5 } & { 82.7 } & { 84.7 } \\
    \midrule
    { XLM-R } & { 86.4 } & { 94.7 } & { 89.7 } & { 90.1 } & { 90.4 } & { 78.7 } & { 79 } & { 82.3 } \\
    \midrule
    { \baseline{} } & { 89.0 } & { 95.9 } & { 91.3 } & { 91.7 } & { 92.4 } & { 82.9 } & { 82.5 } & { 86.4 } \\ 
    \midrule
    { \our{} } & { 90.4 } & { 96.5 } & { 92.7 } & { 93.2 } & { 93.6 } & { 84.6 } & { 84.6 } & { 87.4 } \\
    \midrule
    \multicolumn{9}{@{}c@{}}{\textbf{XL} } \\
    \midrule
    { mT5 } & { 89.6 } & { 96 } & { 92.8 } & { 92.7 } & { 92.4 } & { 83.6 } & { 83.1 } & { 86.5 } \\
    \midrule
    { \baseline{} } & { 90.3 } & { 95.9 } & { 93.2 } & { 93.1 } & { 92.9 } & { 84.8 } & { 84.7 } & { 87.4 } \\
    \midrule
    { \our{} } & { 91.0 } & { 95.9 } & { 92.7 } & { 93.2 } & { 93.7 } & { 86.9 } & { 87.0 } & { 87.8 } \\
    \midrule
    \multicolumn{9}{@{}c@{}}{\textbf{XXL} } \\
    \midrule
    { mT5 } & { 90 } & { 96.3 } & { 92.9 } & { 92.6 } & { 92.7 } & { 84.5 } & { 83.9 } & { 87.2 } \\
    \midrule
    \midrule
    \multicolumn{9}{@{}l}{\textit{Translate-Train}} \\
    \midrule
    \midrule
    \multicolumn{9}{@{}c@{}}{\textbf{Base} } \\
    \midrule
    { mT5 } & { 89.3 } & { 95.5 } & { 90.9 } & { 91.4 } & { 92.5 } & { 83.6 } & { 84.8 } & { 86.4 } \\
    \midrule
    { \baseline{} } & { 91.1 } & { 95.7 } & { 93.1 } & { 92.8 } & { 93.3 } & { 86.6 } & { 87.8 } & { 88.7 } \\
    \midrule
    
    { \our{} } & { 91.8 } & { 96.2 } & { 93.6 } & { 93.6 } & { 94.2 } & { 87.2 } & { 89 } & { 89.1 } \\
    \midrule
    \multicolumn{9}{@{}c@{}}{\textbf{Large} } \\
    \midrule
    { mT5 } & { 91.2 } & { 96.4 } & { 92.7 } & { 93.3 } & { 93.6 } & { 86.5 } & { 87.4 } & { 88.4 } \\
    \midrule
    { \baseline{} } & { 91.9 } & { 96.0 } & { 93.6 } & { 93.4 } & { 94.2 } & { 87.8 } & { 89.2 } & { 89.0 } \\ 
    \midrule
    { \our{} } & { 92.4 } & { 96.7 } & { 94.9 } & { 94.1 } & { 94.3 } & { 87.3 } & { 89.7 } & { 89.5 } \\
    \midrule
    \multicolumn{9}{@{}c@{}}{\textbf{XL} } \\
    \midrule
    { mT5 } & { 91.0 } & { 96.4 } & { 92.5 } & { 93.1 } & { 93.6 } & { 85.5 } & { 86.9 } & { 89.0 } \\
    \midrule
    { \baseline{} } & { 92.2 } & { 96.1 } & { 93.9 } & { 93.6 } & { 94.9 } & { 88.1 } & { 89.4 } & { 89.3 } \\
    \midrule
    { \our{} } & { 92.6 } & { 97.1 } & { 94.2 } & { 94.6 } & { 95.3 } & { 88.4 } & { 88.8 } & { 89.8 } \\
    \midrule
    \multicolumn{9}{@{}c@{}}{\textbf{XXL} } \\
    \midrule
    { mT5 } & { 91.5 } & { 96.1 } & { 92.9 } & { 93.6 } & { 94.2 } & { 87 } & { 87.9 } & { 89.0 } \\
    \bottomrule
    \end{tabular}
    \caption{PAWS-X accuracy scores for each language}
    
\end{table}
\onecolumn

\begin{table*}[!thb]
    \footnotesize
    \centering
    \aboverulesep=0ex
    \belowrulesep=0ex
    \renewcommand{\arraystretch}{1.2}
    \begin{tabular}{@{}l@{}|@{}c@{}|@{}c@{}|@{}c@{}@{}c@{}@{}c@{}@{}c@{}@{}c@{}@{}c@{}@{}c@{}@{}c@{}@{}c@{}@{}c@{}@{}c@{}@{}c@{}@{}c@{}@{}c@{}@{}c@{}@{}r@{}}
    \toprule
    \textbf{{ Model }} & \textbf{{ \# Params }} & \textbf{{ Avg }} & \textbf{{ en }} & \textbf{{ fr }} & \textbf{{ es }} & \textbf{{ de }} & \textbf{{ el }} & \textbf{{ bg }} & \textbf{{ ru }} & \textbf{{ tr }} & \textbf{{ ar }} & \textbf{{ vi }} & \textbf{{ th }} & \textbf{{ zh }} & \textbf{{ hi }} & \textbf{{ sw }} & \textbf{{ ur }} \\
    \midrule
    \midrule
    \multicolumn{18}{@{}l@{}}{\textit{Cross-lingual zero-shot transfer (models fine-tune on English data only)}} \\
    \midrule
    \midrule
    \multicolumn{18}{@{}c@{}}{\textbf{Base}} \\
    \midrule
    mT5 & { 580M } & { 75.4 } & { 84.7 } & { 79.1 } & { 80.3 } & { 77.4 } & { 77.1 } & { 78.6 } & { 77.1 } & { 72.8 } & { 73.3 } & { 74.2 } & { 73.2 } & { 74.1 } & { 70.8 } & { 69.4 } & { 68.3 } \\
    \midrule
    XLM-R & { 225M } & { 76.2 } & { 85.8 } & { 79.7 } & { 80.7 } & { 78.7 } & { 77.5 } & { 79.6 } & { 78.1 } & { 74.2 } & { 73.8 } & { 76.5 } & { 74.6 } & { 76.7 } & { 72.4 } & { 66.5 } & { 68.3 } \\
    \midrule
    \baseline{} & { 477M } & { 78.1 } & { 87.3 } & { 81.9 } & { 82.4 } & { 81 } & { 80.2 } & { 81.1 } & { 79.7 } & { 77.7 } & { 76.4 } & { 78.5 } & { 76.2 } & { 79.0 } & { 72.7 } & { 69.6 } & { 68.3 } \\
    \midrule
    \our{} mCCA & { 477M } & { 79.5 } & { 87.8 } & { 82.9 } & { 83.8 } & { 81.5 } & { 81.7 } & { 81.8 } & { 80.8 } & { 79.1 } & { 79.1 } & { 79.8 } & { 77.7 } & { 79.3 } & { 74.6 } & { 72.7 } & { 69.9 } \\
    \midrule
    \makecell[l]{\our{} DRO +\\ CCM} & { 477M } & { 79.7 } & { 87.3 } & { 82.2 } & { 83.7 } & { 82.6 } & { 82.0 } & { 82.5 } & { 80.2 } & { 78.5 } & { 79.0 } & { 80.1 } & { 77.0 } & { 80.3 } & { 74.4 } & { 74.9 } & { 70.2 } \\
    \midrule
    \our{} CCM & { 477M } & { 80.5 } & { 87.7 } & { 83.7 } & { 84.7 } & { 83.7 } & { 82 } & { 83 } & { 81.5 } & { 79.3 } & { 79.7 } & { 80.3 } & { 77.9 } & { 80.2 } & { 76.1 } & { 75.5 } & { 71.6 } \\
    \midrule
    \multicolumn{18}{@{}c@{}}{\textbf{Large}} \\
    \midrule
    XLM-R\textsubscript{Large} & { 550M } & { 80.9 } & { 89.1 } & { 84.1 } & { 85.1 } & { 83.9 } & { 82.9 } & { 84 } & { 81.2 } & { 79.6 } & { 79.8 } & { 80.8 } & { 78.1 } & { 80.2 } & { 76.9 } & { 73.9 } & { 73.8 } \\
    \midrule
    mT5\textsubscript{Large} & { 1.2B } & { 81.1 } & { 89.4 } & { 84.1 } & { 84.2 } & { 83.4 } & { 83.2 } & { 84.1 } & { 81.5 } & { 80.1 } & { 79.8 } & { 81 } & { 79.4 } & { 80.3 } & { 77.6 } & { 75.4 } & { 73.5 } \\
    \midrule
    { \baseline{}\textsubscript{Large} } & { 840M } & { 81.3 } & { 89.4 } & { 84.7 } & { 85.5 } & { 84.4 } & { 83.5 } & { 84.1 } & { 81.9 } & { 81.3 } & { 80.7 } & { 81.2 } & { 79.2 } & { 81.5 } & { 76.5 } & { 74.1 } & { 72.4 } \\
    \midrule
    \our{}\textsubscript{Large} & { 814M } & { 83 } & { 90.1 } & { 86 } & { 86.7 } & { 85.4 } & { 85.7 } & { 85.3 } & { 83.2 } & { 82.6 } & { 83.4 } & { 82.8 } & { 81.0 } & { 82.5 } & { 78.3 } & { 78.1 } & { 74.3 } \\
    \midrule
    \multicolumn{18}{@{}c@{}}{\textbf{XL}} \\
    \midrule
    XLM-R\textsubscript{XL} & { 3.5B } & { 82.3 } & { 90.7 } & { 85.5 } & { 86.5 } & { 84.6 } & { 84 } & { 85.2 } & { 82.7 } & { 81.7 } & { 81.6 } & { 82.4 } & { 79.4 } & { 81.7 } & { 78.5 } & { 75.3 } & { 74.3 } \\
    \midrule
    mT5\textsubscript{XL} & { 3.7B } & { 82.9 } & { 90.6 } & { 85.3 } & { 81.3 } & { 85.8 } & { 85.4 } & { 85.4 } & { 83.7 } & { 82 } & { 82.2 } & { 81.8 } & { 80.9 } & { 82.7 } & { 80.4 } & { 78.6 } & { 77.0 } \\
    \midrule
    \baseline{}\textsubscript{XL} & { 2.2B } & { 83.7 } & { 91.3 } & { 86.8 } & { 87.4 } & { 86.7 } & { 85.8 } & { 85.9 } & { 84.2 } & { 83.4 } & { 82.7 } & { 83.4 } & { 80.9 } & { 83.1 } & { 80.2 } & { 77.6 } & { 75.7 } \\
    \midrule
    \our{}\textsubscript{XL} & { 2.1B } & { 84.8 } & { 92.2 } & { 87.4 } & { 88.7 } & { 87.3 } & { 87.2 } & { 87.3 } & { 83.8 } & { 84 } & { 84.6 } & { 85.1 } & { 81.9 } & { 83.9 } & { 81.6 } & { 80.5 } & { 77.0 } \\
    \midrule
    \multicolumn{18}{@{}c@{}}{\textbf{XXL}} \\
    \midrule
    XLM-R\textsubscript{XXL} & { 10.7B } & { 83.1 } & { 91.6 } & { 86.2 } & { 87.3 } & { 87 } & { 85.1 } & { 85.7 } & { 82.5 } & { 82 } & { 82.5 } & { 83 } & { 79.5 } & { 82.6 } & { 79.8 } & { 76.2 } & { 74.9 } \\
    \midrule
    mT5\textsubscript{XXL} & { 13B } & { 85.0 } & { 91.6 } & { 86.9 } & { 87.8 } & { 87.3 } & { 87.3 } & { 87.7 } & { 85.1 } & { 83.8 } & { 84.5 } & { 79.8 } & { 81.7 } & { 83.6 } & { 83.2 } & { 80.3 } & { 84.6 } \\
    \midrule
    \midrule
    \multicolumn{18}{@{}l@{}}{\textit{Translate-train (models fine-tune on English training data plus translations in all target languages)}} \\
    \midrule
    \midrule
    \multicolumn{18}{@{}c@{}}{\textbf{Base}} \\
    \midrule
   mT5\textsubscript{Base} & { 300M } & { 75.9 } & { 82 } & { 77.9 } & { 79.1 } & { 77.7 } & { 78.1 } & { 78.5 } & { 76.5 } & { 74.8 } & { 74.4 } & { 74.5 } & { 75 } & { 76 } & { 72.2 } & { 71.5 } & { 70.4 } \\
   \midrule
   XLM-R\textsubscript{Base} & { 225M } & { 79.1 } & { 85.4 } & { 81.4 } & { 82.2 } & { 80.3 } & { 80.4 } & { 81.3 } & { 79.7 } & { 78.6 } & { 77.3 } & { 79.7 } & { 77.9 } & { 80.2 } & { 76.1 } & { 73.1 } & { 73.0 } \\
    \midrule
    \baseline{}\textsubscript{Base} & { 477M } & { 81.7 } & { 88.2 } & { 83.8 } & { 84.7 } & { 83.9 } & { 83.5 } & { 84.1 } & { 82.6 } & { 81.6 } & { 81.1 } & { 82.6 } & { 81.0 } & { 82.5 } & { 77.8 } & { 75.2 } & { 73.7 } \\
    \midrule
    \makecell[lc]{\our{}\\mCCA\textsubscript{Base}} & { 477M } & { 82.4 } & { 88.0 } & { 84.7 } & { 85.6 } & { 84.2 } & { 83.8 } & { 84.4 } & { 83.3 } & { 82.1 } & { 82.2 } & { 82.7 } & { 81.4 } & { 82.9 } & { 79.4 } & { 77.3 } & { 73.3 } \\
    \midrule
    \makecell[lc]{\our{} CCM\textsubscript{Base}} & { 477M } & { 82.9 } & { 88.7 } & { 85.6 } & { 86.1 } & { 85.3 } & { 85.2 } & { 85.8 } & { 83.1 } & { 83.1 } & { 82.9 } & { 83.3 } & { 81.0 } & { 83.7 } & { 79.6 } & { 78.1 } & { 72.7 } \\
    \midrule
    \multicolumn{18}{@{}c@{}}{\textbf{Large}} \\
    \midrule
    mT5\textsubscript{Large} & { 1.2B } & { 81.8 } & { 88.3 } & { 83.8 } & { 84.9 } & { 84.0 } & { 83.7 } & { 84.1 } & { 82.0 } & { 81.0 } & { 80.3 } & { 81.3 } & { 79.9 } & { 81.7 } & { 79.8 } & { 76.4 } & { 75.9 } \\
    \midrule
    XLM-R\textsubscript{Large} & { 550M } & { 83.6 } & { 89.1 } & { 85.1 } & { 86.6 } & { 85.7 } & { 85.3 } & { 85.9 } & { 83.5 } & { 83.2 } & { 83.1 } & { 83.7 } & { 81.5 } & { 83.7 } & { 81.6 } & { 78 } & { 78.1 } \\
    \midrule
    \baseline{}\textsubscript{Large} & { 840M } & { 84.1 } & { 90.1 } & { 86.8 } & { 87.1 } & { 86.0 } & { 86.1 } & { 86.4 } & { 84.8 } & { 83.5 } & { 83.7 } & { 84.4 } & { 81.9 } & { 84.9 } & { 81.2 } & { 78.5 } & { 76.4 } \\
    \midrule
    \our{}\textsubscript{Large} & { 814M } & { 84.9 } & { 90.2 } & { 87.4 } & { 87.9 } & { 86.7 } & { 87.0 } & { 87.4 } & { 85.0 } & { 84.7 } & { 84.8 } & { 85.0 } & { 83.4 } & { 85.0 } & { 82.0 } & { 80.9 } & { 75.9 } \\
    \midrule
    \multicolumn{18}{@{}c@{}}{\textbf{XL}} \\
    \midrule
    mT5\textsubscript{XL} & { 3.7B } & { 84.8 } & { 90.9 } & { 86.8 } & { 87.4 } & { 86.8 } & { 86.4 } & { 86.8 } & { 84.9 } & { 84.4 } & { 84.2 } & { 83.9 } & { 82.3 } & { 84 } & { 83.1 } & { 81.3 } & { 79.4 } \\
    \midrule
    XLM-R\textsubscript{XL} & { 3.5B } & { 85.4 } & { 91.1 } & { 87.2 } & { 88.1 } & { 87 } & { 87.4 } & { 87.8 } & { 85.3 } & { 85.2 } & { 85.3 } & { 86.2 } & { 83.8 } & { 85.3 } & { 83.1 } & { 79.8 } & { 78.2 } \\
    \midrule
    \baseline{}\textsubscript{XL} & { 2.2B } & { 85.5 } & { 90.9 } & { 87.4 } & { 88.3 } & { 87.4 } & { 87.2 } & { 87.6 } & { 85.1 } & { 85.1 } & { 85.1 } & { 86.1 } & { 83.7 } & { 85.4 } & { 82.5 } & { 81.3 } & { 78.9 } \\
    \midrule
    \our{}\textsubscript{XL} & { 2.1B } & { 87.1 } & { 92.2 } & { 88.9 } & { 89.7 } & { 89.1 } & { 89.1 } & { 89.1 } & { 86.2 } & { 86.8 } & { 87.0 } & { 87.3 } & { 85.2 } & { 86.7 } & { 84.5 } & { 83.2 } & { 80.8 } \\
    \midrule
    \multicolumn{18}{@{}c@{}}{\textbf{XXL}} \\
    \midrule
    XLM-R\textsubscript{XXL} & { 10.7B } & { 86.0 } & { 91.5 } & { 87.6 } & { 88.7 } & { 87.8 } & { 87.4 } & { 88.2 } & { 85.6 } & { 85.1 } & { 85.8 } & { 86.3 } & { 83.9 } & { 85.6 } & { 84.6 } & { 81.7 } & { 80.6 } \\
    \midrule
    mT5\textsubscript{XXL} & { 13B } & { 87.8 } & { 92.7 } & { 89.1 } & { 90 } & { 89.8 } & { 89.5 } & { 89.4 } & { 87.6 } & { 87.1 } & { 87.2 } & { 87.5 } & { 85.6 } & { 86.5 } & { 86.5 } & { 84.3 } & { 83.8 } \\
    \bottomrule
    \end{tabular}
    \caption{XNLI accuracy scores for each language\label{tab:xnli-main-all}}
    
\end{table*}
\clearpage

\begin{table*}[!htb]
    \footnotesize
    \centering
    
    \aboverulesep=0ex
    \belowrulesep=0ex
    \renewcommand{\arraystretch}{1.2}
    \begin{tabular}{@{}l@{}|@{}c@{}@{}c@{}@{}c@{}@{}c@{}@{}c@{}@{}c@{}@{}c@{}|@{}r@{}}
    \toprule
    { \textbf{Model} } & { \textbf{en} } & { \textbf{ar} } & { \textbf{de} } & { \textbf{es} } & { \textbf{hi} } & { \textbf{vi} } & { \textbf{zh} } & { \textbf{Avg} } \\
    \midrule
    \multicolumn{9}{@{}c@{}}{\textbf{Base} } \\
    \midrule
    { mT5 } & { 81.7/66.9 } & { 57.1/36.9 } & { 62.1/43.2 } & { 67.1/47.2 } & { 55.4/37.9 } & { 65.9/44.1 } & { 61.6/38.6 } & { 64.4/45.0 } \\
    \midrule
    { \baseline{} } & { 82.1/69.2 } & { 62.4/42.4 } & { 65.7/50.7 } & { 71.2/53.1 } & { 65.12/47.5 } & { 69.8/48.8 } & { 64.6/41.5 } & { 68.7/50.5 }  \\
    \midrule
    { \our{} } & { 83.1/70.3 } & { 63.9/43.9 } & { 68.9/54.0 } & { 73.3/55.1 } & { 69.0/51.7 } & { 72.7/52.0 } & { 68.0/45.2 } & { 71.3/53.2 } \\
    \midrule
    \multicolumn{9}{@{}c@{}}{\textbf{Large} } \\
    \midrule
        { XLM-R } & { 80.6/67.8 } & { 63.1/43.5 } & { 68.5/53.6 } & { 74.1/56.0 } & { 69.2/51.6 } & { 71.3/50.9 } & { 68.0/45.4 } & { 70.7/52.7 } \\
    \midrule
    { mT5 } & { 84.9/70.7 } & { 65.3/44.6 } & { 68.9/51.8 } & { 73.5/54.1 } & { 66.9/47.7 } & { 72.5/50.7 } & { 66.2/42.0 } & { 71.2/51.7 } \\
    \midrule
    { \baseline{} } & { 84.1/71.2 } & { 66.6/46.3 } & { 70.0/54.8 } & { 74.7/56.8 } & { 71.0/53.3 } & { 74.6/53.6 } & { 68.8/44.9 } & { 72.8/54.4 } \\
    \midrule
    { \our{} } & { 85.0/72.3 } & { 68.0/47.6 } & { 72.1/56.9 } & { 75.4/57.1 } & { 72.9/54.7 } & { 75.4/54.0 } & { 71.2/47.6 } & { 74.3/55.7 } \\
    \midrule
    \multicolumn{9}{@{}c@{}}{\textbf{XL} } \\
    \midrule
    { mT5 } & { 85.5/71.9 } & { 68.0/47.4 } & { 70.5/54.4 } & { 75.2/56.3 } & { 70.5/51.0 } & { 74.2/52.8 } & { 70.5/47.2 } & { 73.5/54.4 } \\
    \midrule
    { XLM-R } & { 85.1/72.6 } & { 66.7/46.2} & { 70.5/55.5 } & { 74.3/56.9 } & { 72.2/54.7 } & { 74.4/52.9 } & { 70.9/48.5 } & { 73.4/55.3 } \\
    \midrule
    { \baseline{} } & { 85.2/72.6 } & { 68.1/47.6 } & { 71.1/56.4 } & { 75.7/57.4 } & { 73.1/55.2 } & { 75.4/53.9 } & { 71.3/47.7 } & { 74.3/55.8 } \\
    \midrule
    { \our{} } & { 85.4/72.4 } & { 69.0/48.5 } & { 73.0/57.7 } & { 76.8/58.6 } & { 75.0/56.5 } & { 76.2/54.7 } & { 72.1/48.6 } & { 75.4/56.7 } \\
    \midrule
    \multicolumn{9}{@{}c@{}}{\textbf{XXL} } \\
    \midrule
    { XLM-R } & { 85.5/72.4 } & { 68.6/48.4 } & { 72.7/57.8 } & { 75.4/57.6 } & { 73.7/55.8 } & { 76.0/55.0 } & { 71.7/48.9 } & { 74.8/56.6 } \\
    \midrule
    { mT5 } & { 86.7/73.5 } & { 70.7/50.4 } & { 74.0/57.8 } & { 76.8/58.4 } & { 75.6/57.3 } & { 76.4/56.0 } & { 71.8/48.8 } & { 76.0/57.4 } \\
    \bottomrule
    \end{tabular}
    \caption{MLQA results (F1/EM) for each language.}
\end{table*}


\begin{table*}[!htb]
    \footnotesize
    \centering
    
    \aboverulesep=0ex
    \belowrulesep=0ex
    \renewcommand{\arraystretch}{1.2}
    \begin{tabular}{@{}l@{}|@{}c@{}@{}c@{}@{}c@{}@{}c@{}@{}c@{}@{}c@{}@{}c@{}@{}c@{}@{}c@{}|@{}r@{}}
    \toprule
    { \textbf{Model} } & { \textbf{en}  } & { \textbf{ar} } & { \textbf{bn} } & { \textbf{fi} } & { \textbf{id} } & { \textbf{ko} } & { \textbf{ru} } & { \textbf{sw} } & { \textbf{te} } & { \textbf{Avg} } \\
    \midrule
    \multicolumn{11}{@{}c@{}}{\textbf{Base} } \\
    \midrule
    { mT5 } & { 71.8/60.9 } & { 67.1/50.4 } & { 40.7/22.1 } & { 67.0/52.2 } & { 71.3/54.5 } & { 49.5/37.7 } & { 54.9/32.6 } & { 60.4/43.9 } & { 40.6/31.1 } & { 58.1/42.8 } \\
    \midrule
    { \baseline{} } & { 71.8/57.7 } & { 68.3/50.0 } & { 60.6/45.1 } & { 68.0/52.6 } & { 73.2/56.1 } & { 53.2/40.2 } & { 63.4/38.4 } & { 64.4/48.1 } & { 41.3/27.2 } & { 62.7/46.2 } \\
    \midrule
    { \our{} } & { 73.4/59.1 } & { 71.6/54.1 } & { 63.7/51.3 } & { 66.5/52.3 } & { 77.0/63.4 } & { 57.2/43.5 } & { 68.0/49.0 } & { 67.3/51.1 } & { 59.4/39.3 } & { 67.1/51.5 } \\
    \midrule
    \multicolumn{11}{@{}c@{}}{\textbf{Large} } \\
    \midrule
    { XLM-R} & { 71.5/56.8 } & { 67.6/40.4 } & { 64.0/47.8 } & { 70.5/53.2 } & { 77.4/61.9 } & { 31.9/10.9 } & { 67.0/42.1 } & { 66.1/48.1 } & { 70.1/43.6 } & { 65.1/45.0 } \\
    \midrule
    { mT5 } & { 71.6/58.9 } & { 60.5/40.4 } & { 42.0/23.9 } & { 64.6/48.8 } & { 67.0/49.2 } & { 47.6/37.3 } & { 58.9/36.8 } & { 65.7/45.3 } & { 41.9/29.7 } & { 57.8/41.2 } \\
    \midrule
    { \baseline{} } & { 74.7/62.0 } & { 75.2/57.1 } & { 72.9/56.6 } & { 69.9/54.9 } & { 78.9/66.7 } & { 61.4/47.8 } & { 68.0/44.9 } & { 72.2/56.7 } & { 72.8/45.6 } & { 71.8/54.7 } \\
    \midrule
    { \our{} } & { 75.6/62.0 } & { 77.0/59.9 } & { 74.6/62.8 } & { 74.0/57.5 } & { 80.7/67.1 } & { 66.4/52.2 } & { 69.5/46.3 } & { 76.0/61.3 } & { 72.2/48.4 } & { 74.0/57.5 } \\
    \midrule
    \multicolumn{11}{@{}c@{}}{\textbf{XL} } \\
    \midrule
    { mT5 } & { 80.3/70.9 } & { 81.7/65.5 } & { 74.5/57.5 } & { 79.4/65.3 } & { 83.5/70.4 } & { 70.0/60.5 } & { 71.6/47.8 } & { 77.3/59.7 } & { 77.9/55.8 } & { 77.4/61.5 } \\
    \midrule
    { \baseline{} } & { 79.1/64.3 } & { 78.2/60.3 } & { 76.9/64.1 } & { 75.0/60.2 } & { 84.4/70.3 } & { 66.7/54.8 } & { 76.4/56.3 } & { 78.3/63.7 } & { 75.6/51.1 } & { 76.7/60.6 } \\
    \midrule
    { \our{} } & { 78.2/64.1 } & { 79.3/60.8 } & { 78.8/67.3 } & { 77.7/63.2 } & { 84.9/70.6 } & { 68.5/56.2 } & { 77.0/57.5 } & { 79.9/66.3 } & { 77.7/53.2 } & { 78.0/62.1 } \\
    \midrule
    \multicolumn{11}{@{}c@{}}{\textbf{XXL} } \\
    \midrule
    { mT5 } & { 83.7/72.5 } & { 82.8/66.0 } & { 80.2/63.7 } & { 83.3/70.2 } & { 85.3/73.3 } & { 76.2/64.1 } & { 76.6/55.8 } & { 81.9/66.1 } & { 79.2/58.7 } & { 81.0/65.6 } \\
    \bottomrule
    \end{tabular}
    \caption{TYDi QA GP results (F1/EM) for each language.}
\end{table*}

\clearpage

\begin{table*}[!htb]
    \footnotesize
    \centering
    
    \aboverulesep=0ex
    \belowrulesep=0ex
    \renewcommand{\arraystretch}{1.2}
    \begin{tabular}{@{}l@{}|@{}c@{}@{}c@{}@{}c@{}@{}c@{}@{}c@{}@{}c@{}@{}c@{}@{}c@{}@{}c@{}@{}c@{}@{}c@{}|@{}r@{}}
    \toprule
    { \textbf{Model} } & { \textbf{en} } & { \textbf{ar} } & { \textbf{de} } & { \textbf{el} } & { \textbf{es} } & { \textbf{hi} } & { \textbf{ru} } & { \textbf{th} } & { \textbf{tr} } & { \textbf{vi} } & { \textbf{zh} } & { \textbf{Avg} } \\
    \midrule
    \multicolumn{13}{@{}c@{}}{\textbf{Base} } \\
    \midrule
    { mT5 } & { 84.6/71.7 } & { 63.8/44.3 } & { 73.8/54.5 } & { 59.6/35.6 } & { 74.8/56.1 } & { 60.3/43.4 } & { 57.8/34.7 } & { 57.6/45.7 } & { 67.9/48.2 } & { 70.7/50.3 } & { 66.1/54.1 } & { 67.0/49.0 } \\
    \midrule
    { \baseline{} } & { 84.9/72.9 } & { 70.5/54.3 } & { 78.9/63.2 } & { 75.6/57.8 } & { 78.4/60.8 } & { 71.2/54.5 } & { 75.9/59.7 } & { 68.7/58.8 } & { 71.6/55.4 } & { 75.9/56.4 } & { 65.5/56.9 } & { 74.3/59.2 } \\
    \midrule
    { \our{} } & { 87.2/76.0 } & { 72.9/56.0 } & { 80.0/64.5 } & { 79.6/63.5 } & { 81.2/63.1 } & { 75.3/59.7} & { 77.7/61.5 } & { 70.9/59.5 } & { 74.0/58.7 } & { 77.4/59.2 } & { 69.0/61. } & { 76.8/62.1 } \\
    \midrule
    \multicolumn{13}{@{}c@{}}{\textbf{Large} } \\
    \midrule
    { XLM-R} & { 86.5/75.7 } & { 68.6/49.0 } & { 80.4/63.4 } & { 79.8/61.7 } & { 82.0/63.9 } & { 76.7/59.7 } & { 80.1/64.3 } & { 74.2/62.8 } & { 75.9/59.3 } & { 79.1/59.0 } & { 59.3/50.0 } & { 76.6/60.8 } \\
    \midrule
    { mT5 } & { 88.4/77.3 } & { 75.2/56.7 } & { 80.0/62.9 } & { 77.5/57.6 } & { 81.8/64.2 } & { 73.4/56.6 } & { 74.7/56.9 } & { 73.4/62.0 } & { 76.5/56.3 } & { 79.4/60.3 } & { 75.9/65.5 } & { 77.8/61.5 } \\
    \midrule
    { \baseline{} } & { 87.1/75.5 } & { 75.1/58.1 } & { 82.1/66.0 } & { 80.9/64.0 } & { 82.5/64.3 } & { 77.5/61.3 } & { 80.3/63.7 } & { 73.4/59.4 } & { 76.8/60.8 } & { 79.2/59.0 } & { 70.5/61.6 } & { 78.7/63.1 } \\
    \midrule
    { \our{} } & { 88.1/77.4 } & { 76.3/59.6 } & { 82.6/67.1 } & { 82.5/65.1 } & { 83.9/66.6 } & { 77.9/61.3 } & { 80.2/63.6 } & { 74.3/63.8 } & { 78.5/62.9 } & { 80.6/61.6 } & { 71.4/64.6 } & { 79.7/64.9 } \\
    \midrule
    \multicolumn{13}{@{}c@{}}{\textbf{XL} } \\
    \midrule
    { mT5 } & { 88.8/78.1 } & { 77.4/60.8 } & { 80.4/63.5 } & { 80.4/61.2 } & { 82.7/64.5 } & { 76.1/60.3 } & { 76.2/58.8 } & { 74.2/62.5 } & { 77.7/58.4 } & { 80.5/60.8 } & { 80.5/71.0 } & { 79.5/63.6 } \\
    \midrule
    { XLM-R } & { 89.5/79.0 } & { 78.4/61.6 } & { 81.3/64.1 } & { 82.3/63.9 } & { 84.6/66.2 } & { 78.8/63.2 } & { 81.5/65.0 } & { 76.0/65.5 } & { 73.9/57.9 } & { 81.7/61.8 } & { 72.3/66.1 } & { 80.0/64.9 } \\
    \midrule
    { \baseline{} } & { 89.1/79.0 } & { 78.5/62.0 } & { 82.4/66.9 } & { 81.8/65.5 } & { 84.3/67.1 } & { 79.3/63.4 } & { 82.2/66.9 } & { 75.4/65.1 } & { 78.3/62.5 } & { 81.5/62.9 } & { 71.6/65.1 } & { 80.4/66.0 } \\
    \midrule
    { \our{} } & { 89.4/79.2 } & { 79.2/62.0 } & { 84.1/68.3 } & { 83.5/66.1 } & { 84.9/66.6 } & { 80.4/64.5 } & { 82.9/67.1 } & { 75.0/61.7 } & { 79.5/64.5 } & { 83.2/64.1 } & { 72.7/65.0 } & { 81.3/66.3 } \\
    \midrule
    \multicolumn{13}{@{}c@{}}{\textbf{XXL} } \\
    \midrule
    { XLM-R } & { 89.3/79.4 } & { 80.1/63.7 } & { 82.7/65.8 } & { 83.4/65.5 } & { 83.8/66.0 } & { 80.7/65.4 } & { 82.4/65.4 } & { 76.6/65.6 } & { 76.8/61.7 } & { 82.2/63.0 } & { 74.1/67.4 } & { 81.1/66.3 } \\
    \midrule
    { mT5 } & { 90.9/80.1 } & { 80.3/62.6 } & { 83.1/65.5 } & { 83.3/65.5 } & { 85.1/68.1 } & { 81.7/65.9 } & { 79.3/63.6 } & { 77.8/66.1 } & { 80.2/60.9 } & { 83.1/63.6 } & { 83.1/73.4 } & { 82.5/66.8 } \\
    \bottomrule
    \end{tabular}
    \caption{XQuAD results (F1/EM) for each language.}
\end{table*}

\section{Hyperparameters for Fine-Tuning}

In Table~\ref{table:hparam}, we report the hyperparameters for fine-tuning \our{} on the downstream tasks.

\begin{table}[!htb]
\centering
\small
\begin{tabular}{lrrrrr}
\toprule
& XQuAD & MLQA & TyDiQA & XNLI & PAWS-X \\ \midrule
Batch size & 32 & 32 & 32 & 32 & 32 \\
Learning rate & \{2,3,4\}e-5 & \{2,3,4\}e-5 & \{2,3,4\}e-5 & \{5,...,8\}e-6 & \{8,9,10,20\}e-6 \\
LR schedule & Linear & Linear & Linear & Linear & Linear \\
Warmup & 10\% & 10\% & 10\% & 12,500 steps & 10\% \\
Weight decay & 0 & 0 & 0 & 0 & 0 \\
Epochs & 4 & \{2,3,4\} & \{10,20,40\} & 10 & 10\\
\bottomrule
\end{tabular}
\caption{Hyperparameters used for fine-tuning on the downstream tasks.}
\label{table:hparam}
\end{table}